\documentclass[conference]{IEEEtran}
\IEEEoverridecommandlockouts
\usepackage{cite}
\usepackage{amsmath,amssymb,amsfonts}
\usepackage{algorithmic}
\usepackage{graphicx}
\usepackage{textcomp}
\usepackage{xcolor}
\usepackage{subfig}
 \usepackage{floatrow}  
\def\BibTeX{{\rm B\kern-.05em{\sc i\kern-.025em b}\kern-.08em
    T\kern-.1667em\lower.7ex\hbox{E}\kern-.125emX}}
\begin{document}

\title{Unsupervised Domain Adaptation with Implicit Pseudo Supervision for Semantic Segmentation
}

\author{Wanyu~Xu,
        Zengmao~Wang,
        Wei~Bian
\thanks{W. Xu, Z. Wang are with the School of Computer Science, Wuhan University, Institute of Artificial Intelligence, Wuhan University, National Engineering Research Center for Multimedia Software, Wuhan University, Wuhan, China, 430072. B. Wei is Lecturer with the school of computer science at UTS, and he is an ARC DECRA Fellow. His research interests include theoretical and applied machine learning, computer vision, and pattern recognition. (email:xuwanyu@whu.edu.cn, wangzengmao@whu.edu.cn, Wei.bian@uts.edu.au) }}

\maketitle

\begin{abstract}
Pseudo-labelling is a popular technique in unsupervised domain adaptation for semantic segmentation. However, pseudo labels are noisy and inevitably have confirmation bias due to the discrepancy between source and target domains and training process. 
In this paper, we train the model by the pseudo labels which are implicitly produced by itself to learn new complementary knowledge about target domain. Specifically, we propose a tri-learning architecture, where every two branches produce the pseudo labels to train the third one. 
And we align the pseudo labels based on the similarity of the probability distributions for each two branches. 
To further implicitly utilize the pseudo labels, we maximize the distances of features for different classes and minimize the distances for the same classes by triplet loss. 
Extensive experiments on GTA5 to Cityscapes and SYNTHIA to Cityscapes tasks show that the proposed method has considerable improvements.

\end{abstract}


\section{Introduction}

Semantic segmentation aims to assign a semantic class label to each pixel of image and is a crucial approach to provide comprehensive scene understanding for various real-world applications, such as self-driving, robots. 
Deep learning\cite{Liu2020,Su2021,Ma2021a}  with large-scale labeled images for supervised learning\cite{Sun2021} may be the most effective approach to achieve high precision of semantic segmentation\cite{Long2015}. However, labeling each pixel in an image by manual labor is extremely expensive. The more complex scene in an image, the harder to label the image. The available training data for semantic segmentation task are extremely limited.
Hence, domain adaptation is extensively explored in machine learning community and computer vision by utilizing the large-scale labeled data in source domain for target domain task to address this challenge. In this work, we focus on the synthetic-to-real project, which predicts real-world unlabeled data with massive synthetic labeled data\cite{Ros2016, Richter2016}. 



Studies on this topic are based on theoretical insights in \cite{Lee2013PseudoLabelT} that pseudo labels, which pick up the class with the maximum predicted probability, are used as true label to serve as entropy minimization. However, pseudo labels are often noisy and have huge confirmation bias due to huge domain gap and training process. For example, the discrepancy between two domains results in the different space structure for the same class and misclassification on target domain. The pseudo labels are usually generated based on existing model, which inevitably can not bring new and useful target-specific knowledge for model.
Therefore, many methods attempt to utilize less but more confident pseudo labels. 
\cite{wang2020, Pan2020,Subhani2020, Luo2018} rely on fixed probability threshold to select part of pseudo-labels. \cite{Zou2018,Zou2019,Mei2020,Guo2021} rectify the per-class or per-pixel probability on the basis of the past predictions to re-calculate pseudo labels. \cite{Tranheden2020} mixes up the source and target images to generate 
Although these methods succeed in improving pseudo labels, the proposed pseudo labels suffer from confirmation bias and big consumption for pixel-level comparison.

To further improve the reliability of pseudo labels, some works attempt to learn different views as several branches in an end-to-end manner. 
\cite{Saito17a} proposes an asymmetric tri-learning approach for domain adaptation on the basis of three identical branches. This approach forces two branches to be learned different from each other and to provide pseudo labels, which select the high prediction probability on the two branches. The other branch is only trained with the pseudo labels for domain-specific knowledge learning. 
In \cite{Zheng2020a}, one primary classifier and one auxiliary classifier with weak constraint are utilized to produce different views. An extra regularization are utilized to keep two classifiers diverse.
\cite{Zhang2021} maintain extra momentum encoder with the same structure to generate pseudo labels. And augmented images serve as an augmented view to measure the reliability of pseudo labels.
Although these methods have achieved impressive results, their main contributions are similar to the self-training approaches, and the ensemble of different views are not well achieved. These methods are easy to accumulate classification bias if the wrong pseudo labels have high prediction probability.

To address the above issues, we propose an unsupervised domain adaptation method with implicit pseudo supervision based on a tri-training architecture for semantic segmentation, termed as EPS-UDA. In the proposed tri-learning architecture, three different segmentation networks follow a shared bottleneck and each two networks provide pseudo labels for the third one.
This form can guarantee that pseudo labels are not directly generated from the trained network and provide new complimentary knowledge for the trained networks continuously.
Implicit pseudo supervision aims to provide reliable target-domain knowledge while keeping diversity without regularization for tri-learning architecture, including semantic feature alignment(SFA) and adaptation ability estimation (AAE).
The proposed SFA aligns the feature centroids based on pseudo labels for each class. Considered the difference between classes, we split the classes into background and foreground category and implement two strategies for different categories. 
The proposed AAE measures how reliable the pseudo labels are generated from two networks and how much these pseudo labels can improve for the architecture. This measurement is based on the similarity between two networks and targeted to explore the adaptation ability for each pixel and each network.
The ensemble of different prediction from the three segmentation networks are used because the three branches are all the same important for the final prediction. 
The experimental results of the proposed method on two popular synthetic-to-real projects in adaptive semantic segmentation tasks outperform most of the state-of-the-art methods considerably. We summarize the main contributions as follows:

\begin{figure}[!t]
\subfloat[self-training method]{\includegraphics[width=\linewidth]{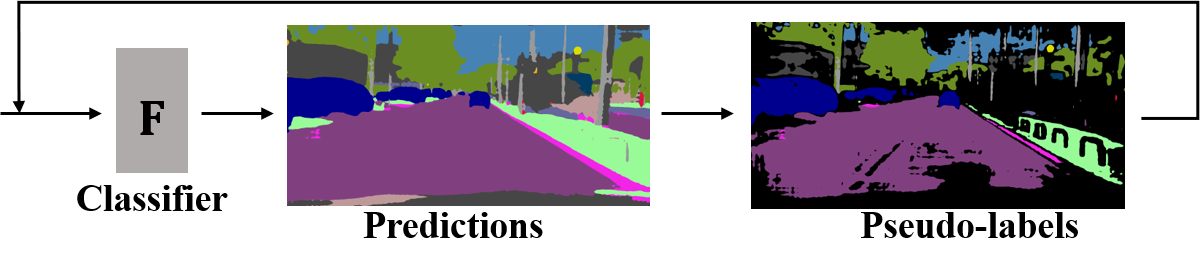}}
\hfil
\subfloat[proposed method]{\includegraphics[width=\linewidth]{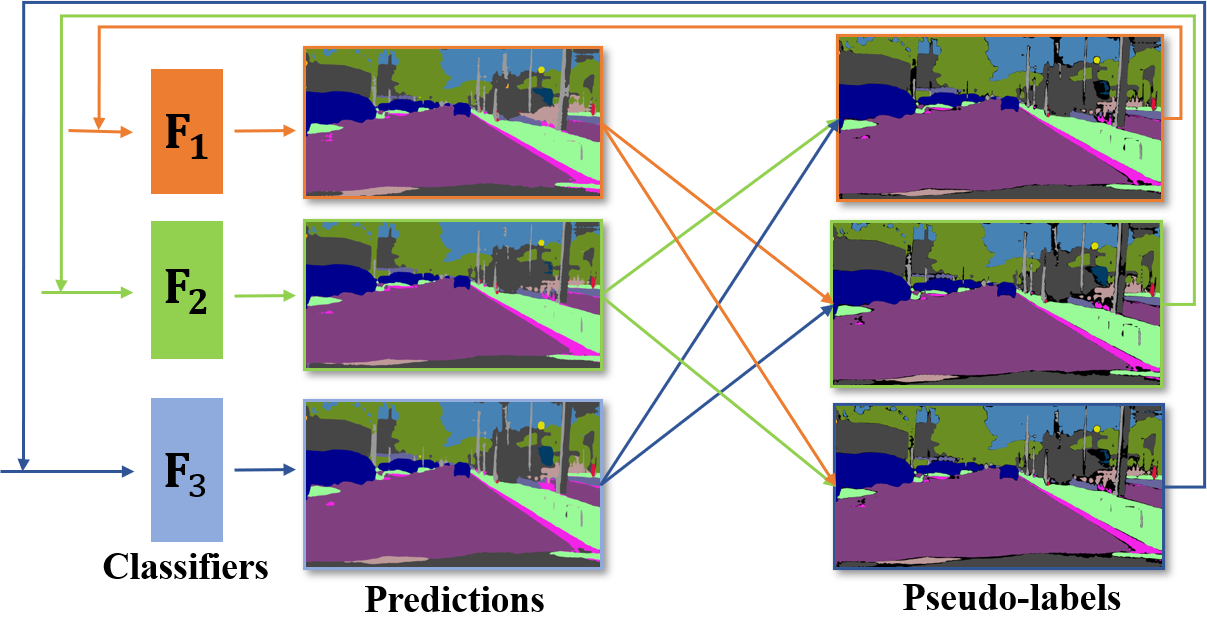}}
\caption{Comparison between self-training method and our proposed method.\textmd{ Normal arrows represent the process of generating pseudo labels, and the dotted arrows represent the back propagation by pseudo labels. Our proposed method can provide complimentary knowledge to each branch and avoid confirmation bias.}}
\label{fig:compare}
\end{figure}

\begin{itemize}
    \item We propose an unsupervised domain adaptation method for semantic segmentation with implicit pseudo supervision. This method is based on tri-learning architecture where each two networks provide pseudo labels for the third one. This form can avoid confirmation bias and provide new complimentary target-domain knowledge continuously in self-training process.
    \item Implicit pseudo supervision aims to provide reliable target-domain knowledge while keeping diversity without regularization for tri-learning architecture. It not only aligns the feature centroids for each class in different strategies, but also rectifies the pseudo labels based on the exploit of tri-learning architecture.
    \item We conduct extensive experiments by using GTA5 and SYNTHIA as the source domain and Cityscapes as the target domain. The improvements of the proposed method are considerable.
\end{itemize}

\section{Related Works}

\subsection{Unsupervised Domain Adaptation for Semantic Segmentation}

The main challenge in unsupervised domain adaptation is the different data distributions between the source domain and the target domain\cite{Long2015,Ros2016,Richter2016,Cordts2016}. Many works have focused on deep learning to address the challenge. Previous deep learning methods can be mainly divided into two categories. 

Without loss of generality, the first category attempts to minimize the distribution discrepancy between source domain and target domain directly, mostly either from image level, e.g. adversarial training directly on the output\cite{Tsai2018, Luo2018, Luo2019a, Yang2020} or image-to-image translation \cite{Murez2017,Kang2018,Yang2020}, either from representation level, e.g. adversarial training directly on the feature layer\cite{wang2020a, Guo2021,Chang2019} or from multi-scale level\cite{Lian2019,Zheng2020,Zhong2020}. Some studies minimize the discrepancy by aligning the distributions explicitly with specific objects function such as Maximum Mean Discrepancy(MMD)\cite{Tsai2018, Ma2021}, Entropy Minimization\cite{Vu2019} and Wasserstein Distance\cite{Lee2019}. 
Although these techniques are effective, the ability to reduce the distribution discrepancy sometimes is unsatisfactory. To further improve the adaptation ability, some studies transfer the source images to target image previously with unsupervised manner, such as the style transfer\cite{Kim2020, Hoffman2017, Li2019a}. 
Recent works\cite{Tranheden2021,Gao2021} try to mix up the images to provide a intermediate domain.
As stated in \cite{Yang2020}, these methods mainly focus on the common knowledge and ignores the private knowledge from a certain domain.

The second category is developed to learn the domain-specific knowledge. It usually consists two steps: 
first, the domain-specific knowledge is learned and generated for target domain. Second, this knowledge is utilized to improve the adaptation model. Self-training\cite{Zheng2020, wang2020a, Zou2018,Zou2019, Li2020}, is a popular method to learn the domain-specific knowledge for target domain and assigns and updates pseudo labels in an alternative style.  
\cite{wang2020a, Zou2018, Zou2019,Zhong2020,Mei2020} normally treat the pseudo labels as true label and \cite{Zhang2021, Pan2019, wang2020a} calculate the centroid of each category to reduce noise. \cite{Li2020} propose to align the category distribution vertically and horizontally. The improvement of pseudo labels has been widely investigated because pseudo-labels are often noisy and unreliable. \cite{wang2020a, Ma2021} select a fixed ratio of the most confident pseudo-labels. \cite{Zou2018,Zou2019} calculate probability threshold for each category from previous probability. \cite{Zheng2020, Guo2021, Zhang2021, Mei2020} rectify the probability for each pixel with adaptive threshold.

Many other methods have attempted to solve the challenge from a novel aspect. \cite{Huang2020,Tsai2019, Zhang2017}  use cluster algorithms first to analyze target data as prior.  \cite{Pan2020,Subhani2020} break the huge gap into small gaps and progressively align the model. \cite{Yang2020a} applies apply Fourier Transformation to align the frequency between two domains. \cite{Kang2020}  maintains consistency in cycle association , that is, the cycle from source to target and to source.

\subsection{Multiview Learning for Adaptive Semantic Segmentation}

Multiview learning aims to train different models with different views of the data. Ideally, these views complement each other, and the models can collaborate in improving each other's performance\cite{Opitz1999,Herrera2016}. As a semi-supervised method, multiview learning has been successfully applied in adaptive semantic segmentation and made a large progress. \cite{Tsai2018, Zheng2020, Zheng2020a} utilize two classifiers, including a main classifier and an auxiliary classifier, to train the same models. The auxiliary classifier helps THE main classifier align the distribution on different feature level. 
\cite{Saito2018, Luo2019a,Luo2018} utilize two same classifiers equally to train the same model and extra loss to maximize the difference of classifiers' weight to keep different views.These methods can implement the pixel-level adaptation easily by comparing the different predictions for each pixel.
\cite{Lian2019, Iqbal2019} utilize multiscale predictions generated from the same feature to force the model to learn different views in one model.
\cite{Zhang2021} uses contrasive learning, which forms different views in one classifier by data augmentation. The enhanced view can be regarded as the teacher view to instruct the model in target domain. \cite{Saito17a} proposes an asymmetric tri-learning architecture for unsupervised domain adaptation. It has three branches with the same structure but are trained asymmetrically. Two auxiliary branches are trained with source and target domains, and their diversity is maintained by maximizing their weight discrepancy. The other branch is trained with only target pseudo labels where a high fixed probability threshold is adopted to select more confident pseudo labels progressively. Hence, obtaining diverse views and reliable pseudo-labels is difficult. This method still suffers from cumulative classification error due to common drawback for self-training techniques.

\section{Methodology}
\begin{figure*}
   \includegraphics[width=\textwidth]{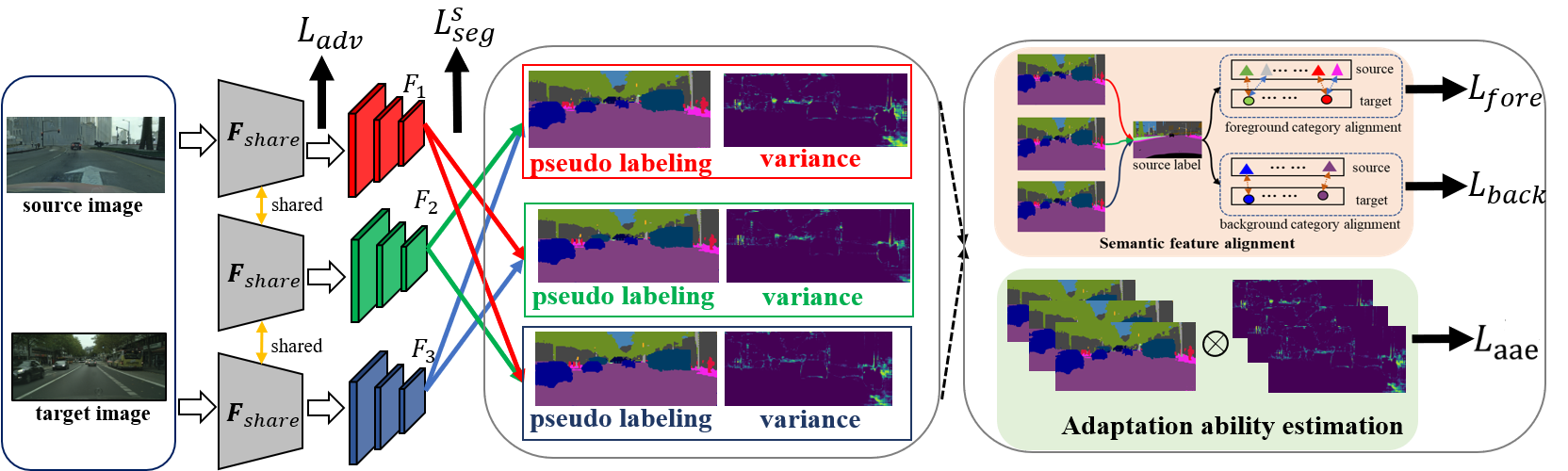}
    \caption{Overview of EPS-UDA. \textmd{
    Tri-learning architecture has a backbone network $F_{share}$ and 
    three segmentation networks $F_1, F_2, F_3$.
    Each network calculates source segmentation loss $\mathcal{L}^s_{seg}$. 
    Target pseudo labels for each segmentation network are generated by two other predictions during pseudo labeling. Implicit pseudo supervision includes semantic feature alignment(SFA) and adaptation ability estimation (AAE). SFA minimizes the distance of feature centroids between the same class for background categories to get $\mathcal{L}_{back}$ and maximize the distance between different categories for foreground categories to get $\mathcal{L}_{fore}$.
    AAE exploit the adaptation ability for each pixel and each network and rectifies the pseudo labels to get $\mathcal{L}_{aae}$.
    } }
  \label{fig:model}
\end{figure*}

In this section, we first introduce the basic framework of adaptive semantic segmentation with adversarial learning. We extend the basic framework to the proposed network architecture, which includes three segmentation networks, denoted as tri-learning architecture. On the basis of the tri-learning architecture, we use any two branches to provide the pseudo-labels for the other branch training. Two implicit pseudo supervision strategies are proposed to align the source domain and target domain by using the pseudo labels and to ensure that each segmentation network is well adapted. The architecture of the proposed method is shown in Fig. \ref{fig:model}.


\subsection{Preliminary}
In a domain adaptation scenario for the semantic segmentation, we denote the source dataset as $X_s = \{ x^s_i, y^s_i \}^{n_s}_{i=1}$, 
target dataset as $X_t = \{ x^t_i \}^{n_t}_{i=1}$ without annotation.
The source cross-entropy loss to train a segmentation network can be represented as


\begin{equation}
\begin{split}
 \mathcal{L}^s_{seg} &= - \sum_{hw}^{HW}  y_{hw}^s \log (F(x_{hw}^s)) 
\end{split}
\end{equation}
where $h,w$ represents the row and column in a image.

The adversarial network is usually adopted to reduce the domain discrepancy between source dataset and target dataset. The adversarial loss can be represented as
\begin{equation}\label{eq:adv}
  \begin{split}
    \mathcal{L}_{adv} = & - \sum_{hw}^{HW} \mathbb{E}[\log(F(x_{hw}^t))] - \sum_{hw}^{HW}\mathbb{E}[\log(1 - F(x_{hw}^{s}))]
  \end{split}
\end{equation}
Hence, the loss to train an adaptive semantic segmentation network can be represented as
\begin{equation}\label{eq:traditional}
\mathcal{L} = \mathcal{L}^s_{seg} + \lambda_{adv}  \mathcal{L}_{adv}
\end{equation}
where $\lambda_{adv}$ is a trade-off parameter.  We can obtain a basic adaptive network for semantic segmentation by optimizing the loss function in Eq. \ref{eq:traditional}.

\subsection{Tri-learning Architecture}
Self-training technique to utilize pseudo labels is a popular technique to utilize the pseudo-labels. However, the self-training is easy to provide wrong pseudo-labels with high probabilities because no extra auxiliary information can be found to rectify these pseudo labels. One of the reasons why the performance of self-training is limited is the confirmation bias between two domains and the model itself.

Inspired by the multi-view strategy
, we propose a tri-learning architecture with three segmentation networks to provide more accurate pseudo labels. In the proposed architecture, three segmentation networks and an adversarial network follows a shared backbone network. We denote the backbone network as $F_{share}$, the three segmentation networks as $F_i, i=1,2,3$ respectively, and the adversarial network as $D$. The source segmentation loss for $F_i$ is noted as $\mathcal{L}^{s,i}_{seg}$. Then the tri-learning architecture can be trained with the following loss

\begin{equation}\label{eq:advp}
  \begin{split}
    \mathcal{L}_{adv} = & - \sum_{hw}^{HW} \mathbb{E}[\log(F_{share}(x_{hw}^t))] \\
        & - \sum_{hw}^{HW}\mathbb{E}[\log(1 - F_{share}(x_{hw}^{s}))] \\
        \mathcal{L}^{s,i}_{seg} &= -\sum_{hw}^{HW} y_{hw}^s log(F_{share}(F_i(x_{hw}^s)) \\
  \end{split}
\end{equation}
On the basis of these predictions, we attempt to assign pseudo labels for the pixels in the target domain. For convenience, given a pixel $x_{hw}^t$, we denote the predictions obtained by $F_i$ as $y_{hw}^{i}$, the pseudo label for $F_i$ as $\hat{y}_{hw}^{i}$. In the training, we use the pseudo labels provided by any two segmentation networks to train the other one.This process can avoid the problem caused by the self-training strategy. 
\begin{equation}\label{eq:pseudo}
  \begin{split}
    \hat{y}^{i}  = \{ y_{hw}^{j} |  y_{hw}^{j} = y_{hw}^{k}, j \neq k, k \neq i, j \neq i \} 
  \end{split}
\end{equation}
The tri-learning architecture is trained with the pseudo labels and can be represented as
\begin{equation}
  \begin{split}
     \mathcal{L}^{t,i}_{seg} &= - \sum_{hw}^{HW} \hat{y}_{hw}^t log(F_{i}(F_{share}(x_{hw}^t)) 
  \end{split}
\end{equation}

The tri-architecture can be optimally trained by using the source images and the pseudo labels from target domain as
\begin{equation}
  \begin{split}
  \mathcal{L} =\lambda_{adv}\mathcal{L}_{adv} + \sum_{i=1}^3 ( \mathcal{L}_{seg}^{s,i} + \mathcal{L}_{seg}^{t,i} )
  \end{split}
\end{equation}

\begin{figure}\label{fig:triplet}
    \centering
    \includegraphics[width=\linewidth]{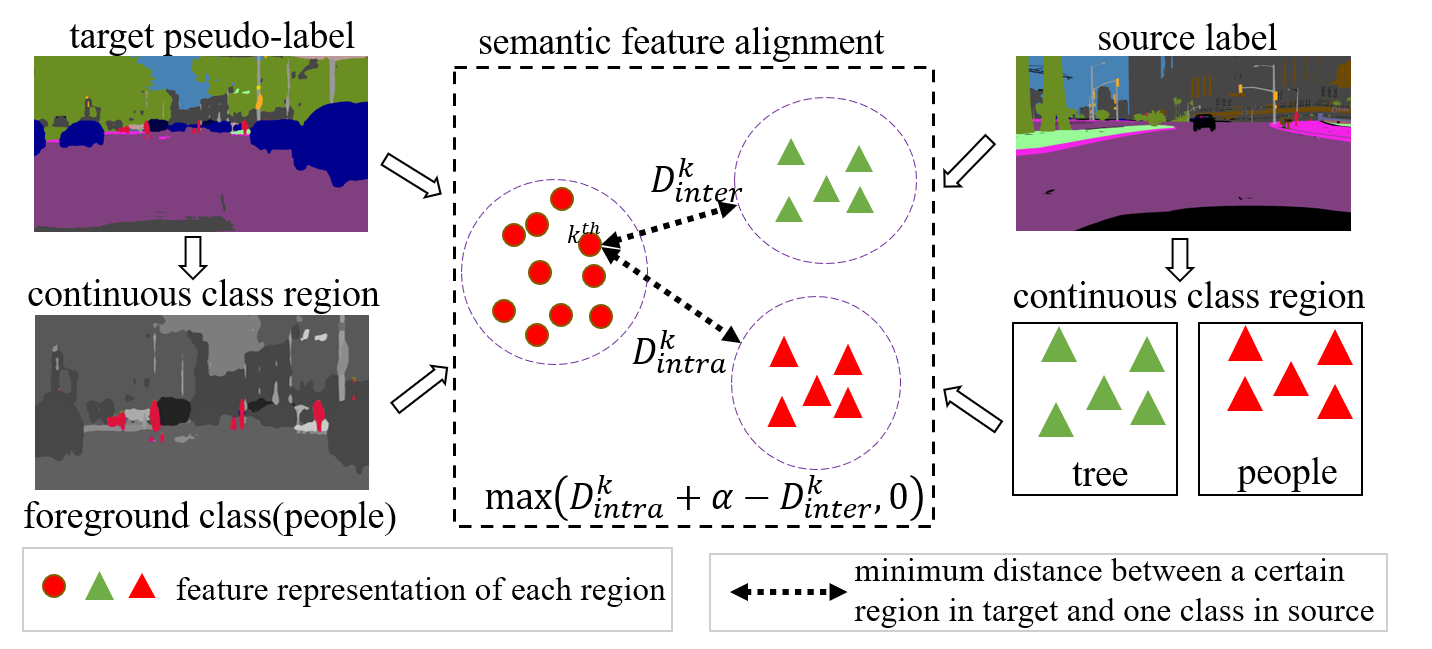}
    \caption{Illustration of alignment of foreground categories in SFA.\textmd{ Take ``people'' class in red color as example. We first extract target features as red circles for every red continuing region in target pseudo-labels. We then align the target features with the all types of extracted source features as green triangles and red triangles, for ``tree'' class and ``people'' class respectively. We conditionally maximize the distances of features in the different categories and minimize the distances in the same categories on the basis of a margin. 
    }}
\end{figure}

\subsection{Implicit Pseudo Supervision}
To fully utilize the pseudo labels, we propose an implicit pseudo supervision including SFA and AAE to align the semantic structures between the target domain and source domain. This supervision not only improve the performance of each segmentation network as self-training does, but also keeps the diversity in tri-learning architecture without extra regularization. 

\subsubsection{\textbf{SFA}}
SFA align the feature centroids for classes conditionally to avoid explicitly aligning the predictions with pseudo labels. 

We divide the categories into the background category $B$ and the foreground category $R$. The background categories have large continuous regions in the image and lack variation for pixels, such as sky and road. Hence, simply shortening the distance of the feature centroids between two domains is effective enough to align the background categories.
The foreground categories may take over a small region in the image and are confusing with other foreground categories, such as traffic sign and traffic pole. Hence, we should consider not only the relationship of the same class, but also relationships of different classes. 

For the background categories, the feature centroid of each category is represented by the average features of all the pixels that belong to the same category. We denote the source feature centroid for segmentation network $F_i$ and class $k$ as $c^{s,i}_{k}$.
\begin{equation}
c^{s,i}_{k} = \frac{\sum_{hw}^{HW} \mathbb{I}[y_{hw}^s = k] * f_{hw}}{\sum_{hw}^{HW} \mathbb{I}[y^s = k] }
\end{equation}
where $\mathbb{I}$ is indicator function. $f_{hw}$ is the feature of a pixel $x_{hw}$. We save the $n_B$ latest calculated source feature centroids as $ \{f^{s,i}_{k,l}\}^{n_B}_{l=1}$ in each iteration. 
With the pseudo labels $\hat{y}^{i}$, we can get the target feature centroid $c^{t,i}_{k}$ in a similar way. Then we can simply shorten the distance between target feature centroid and closest source feature centroid.
\begin{equation}
    \mathcal{L}_{back}^i = \sum_{k \in B} \min_{l \in n_B} | c^{t,i}_{k} - c_{k,l}^{s,i} |
\end{equation}


For the foreground categories, 
we firstly extract top $M$ largest connected area for segmentation network $F_i$ and class $k$ noted as $A_k^{i} = \{ a_{m} \}^{M}_{m=1}$. Then the source feature centroid $c^{s,i}_{k,m}$ can be represented as
\begin{equation}
c^{s,i}_{k,m} = \frac{\sum_{hw}^{HW} \mathbb{I}[x_{hw}^s \in a_m ] * \mathbb{I}[y_{hw}^s = k] * f_{hw}}{\sum_{hw}^{HW} \mathbb{I}[y_{hw}^s \in a_m ] * \mathbb{I}[y^s = k] }
\end{equation}
Like the strategy for background categories, we save the $n_R$ latest calculated source feature centroids as $ \{c^{s,i}_{k,l}\}^{n_R}_{l=1}$ in each iteration. And we calculate the target feature centroid $c^{t,i}_{k,m}$ with pseudo labels. 
Unlike the strategy for background categories, we firstly calculate the closest distance of feature centroids between two domains as $D_{intra}^{i,k_1}$ for class $k_1$ and segmentation network $F_i$. Then we calculate the closest distance of centroids for two different classes $k_1$ and $k_2$ and network $F_i$ as $D_{inter}^{i,k_1,k_2}$. Finally, we can assume the distance of centroids for the same classes should be shorter than the that for different classes to an extent $\alpha$.

\begin{equation}\label{eq:triplet}
\begin{split}
& D_{intra}^{i,k_1} = \sum^{M}_{m=1} \min_{l \in n_R } | c^{t,i}_{k_1,m} - c^{s,i}_{k_1,l} | \\
& D_{inter}^{i,k_1,k_2} = \sum^{M}_{m=1} \min_{l \in n_R } | c^{t,i}_{k_1,m} - c^{s,i}_{k_2,l} | \\
 \mathcal{L}^i_{fore} = \sum_{k_1 \in R } & \quad \sum_{k_2 \in R, k_2 \neq k_1} \max ( D_{intra}^{i,k_1} - D_{inter}^{i,k_1,k_2} + \alpha , 0 )
\end{split}
\end{equation}

So the total SFA loss is to add background category loss and foreground category loss.
\begin{equation}
\begin{split}
  \mathcal{L}^i_{sfa} = \sum_{i=1}^3 ( \mathcal{L}^i_{fore}  + \mathcal{L}^i_{back} )
\end{split}
\end{equation}

\subsubsection{\textbf{AAE}}

AAE measures how much a pseudo-labelled pixel can improve the model and adaptively rectifies the pseudo labels so that the model can learn the complimentary domain-specific knowledge.

Given the probability distribution $p^{i}$, the pseudo labels $\hat{y}^i$ for segmentation network $F_i$ and target domain, we can calculate the average probability distribution $\overline{p}^{i}$ from the probability distributions of other segmentation networks $F_{j}, F_{k}$.  Then we can evaluate the confidence of pseudo label as $\mathcal{M}^i$ base on multi-view learning, i.e., the more differently probability distributions agree, the more confident the prediction is. Finally, we re-align the pseudo labels for target domain to explore the target-specific knowledge. 
\begin{equation}\label{eq:similarity}
\begin{split}
     \overline{p}^i &= ( p^{j} + p^{k} ) / 2  \\
     \mathcal{M}^{i}  &= - (D_{KL}(p^{j}| \overline{p}^i) + D_{KL}(p^{k} | \overline{p}^i) ) \\
    \mathcal{L}^{i}_{aae}=& - \sum_{hw}^{HW}\mathcal{M}^{i}*\hat{y}^t log(F_{share}(F_i(x_{hw}^t))
\end{split}
\end{equation}
where $j \neq i, j \neq k, k \neq i$, $D_{KL}$ represent KL-divergence which is common to calculate the similarity between two distribution.

\begin{figure}
     \centering
     \includegraphics[width=\linewidth]{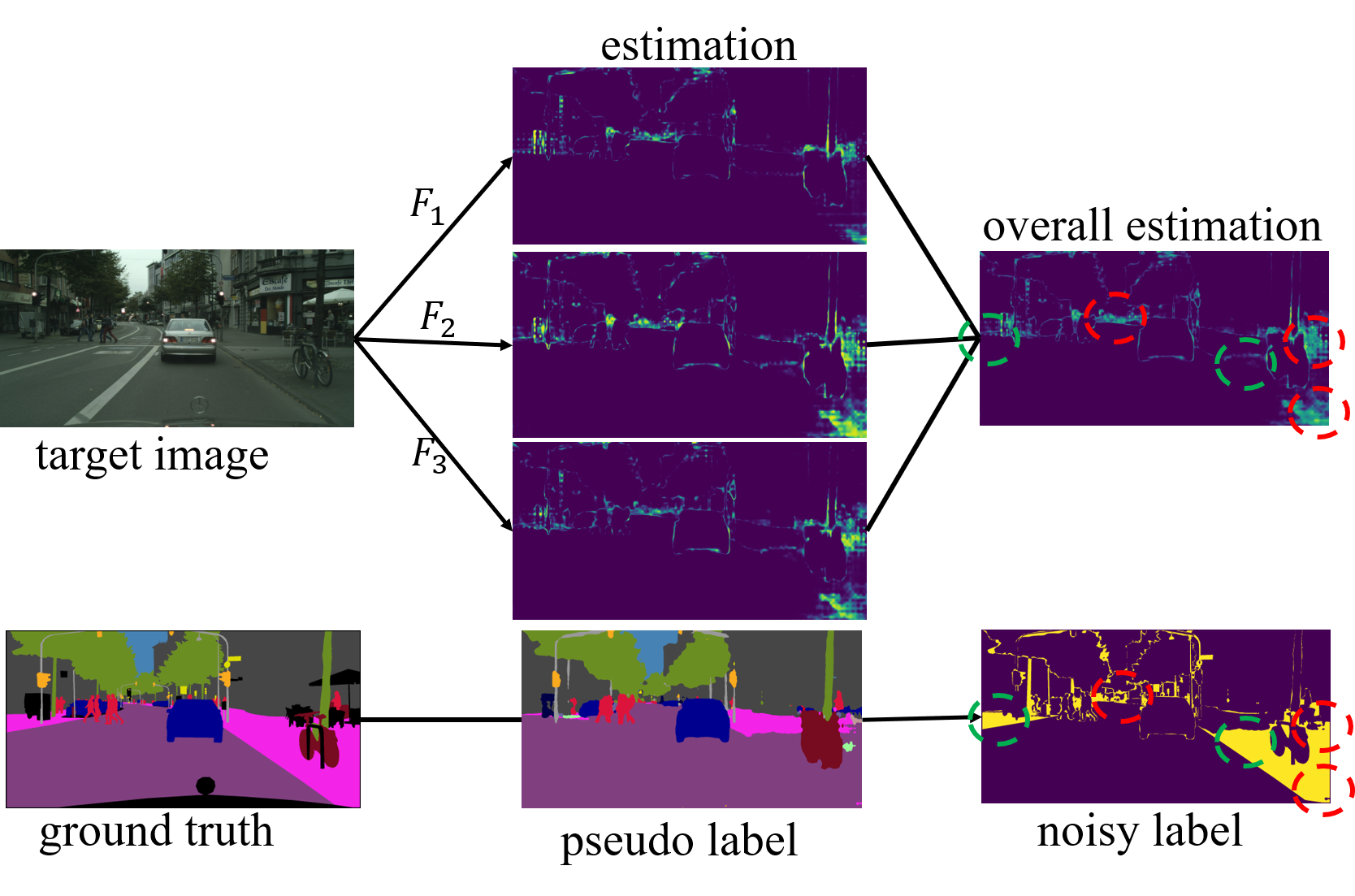}
 \caption{Illustration of AAE. The variance between two predictions that generate pseudo-labels for each branch is showed as an estimation. The overall estimation is the sum of three estimations.
 In the estimation,the cyan area represents the high variance, and the purple area represents the low variance. In the noisy label,the yellow area represents the inconsistency between ground truth and pseudo labels, and the purple area represents the consistency.}
     \label{fig:estimate}
 \end{figure}

To understand the function of AAE more clearly, we show an example to introduce the AAE in Fig \ref{fig:estimate}.
The high-value areas in the estimation (inside the red circle) are often consistent with the staggered ares in noisy label (inside the red circle), which are mixed with correct and wrong pseudo labels. The low-value ares in estimation (inside green circle) cover the pure yellow areas (inside green circle), which are full of wrong pseudo labels. Thus, the high-value areas represent poorly aligned areas that need to be forced to align. The low-value areas represent well aligned areas regardless of how wrong or how correct the pseudo labels are and should be slightly aligned .

\
\subsubsection{Summary} 
The final loss for the proposed method can be represented as
\begin{equation}
\mathcal{L} =  \lambda_{adv} \mathcal{L}_{adv} + \sum_{i=1}^3 ( \mathcal{L}_{seg}^{s,i} + \lambda_{aae}  \mathcal{L}_{aae}^{i} + \lambda_{sfa}(\mathcal{L}_{fore}^i + \mathcal{L}_{back}^i ) ) 
\end{equation}

\section{Experiments}
\begin{table*} 
\resizebox{\textwidth}{!}{
 \centering
  \caption{Comparison between EPS-UDA and other state-of-the-art methods on GTA5 $\to$ Cityscapes }
  \label{tab:gta5}
    \setlength{\tabcolsep}{0.007\textwidth}
\begin{tabular}{l| ccc ccc ccc ccc ccc ccc c|c}
      \hline
    method & road & S.W. & build & wall &  fence & pole & light & sign & Veg. & Ter. & sky & P.R. & rider & car & truck & bus & train & motor & bike & mIoU\\
    \hline
Source\cite{Tsai2018} &  75.8 & 16.8 & 77.2 & 12.5 & 21.0 & 25.5 & 30.1 & 20.1 & 81.3 & 24.6 & 70.3 & 53.8 & 26.4 & 49.9 & 17.2 & 25.9 & 6.5 & 25.3 & 36.0 & 36.6 \\
Adapt\cite{Tsai2018}  & 86.5 & 36.0 & 79.9 & 23.4 & 23.3 & 23.9 & 35.2 & 14.8 & 83.4 & 33.3 & 75.6 & 58.5 & 27.6 & 73.7 & 32.5 & 35.4 & 3.9 & 30.1 & 28.1 & 42.4 \\
CrCDA\cite{Huang2020} & 92.4 & 55.3 & 82.3 & 31.2 & 29.1 & 32.5 & 33.2 & 35.6 & 83.5 & 34.8 & 84.2 & 58.9 & 32.2 & 84.7 & 40.6 & 46.1 & 2.1 & 31.1 & 32.7 & 48.6 \\
SIM\cite{wang2020} & 90.6 & 44.7 & 84.8 & 34.3 & 28.7 & 31.6 & 35.0 & 37.6 & 84.7 & 43.3 & 85.3 & 57.0 & 31.5 & 83.8 & 42.6 & 48.5 & 1.9 & 30.4 & 39.0 & 49.2\\
BCDM\cite{Li2020b} & 90.5 & 37.3 & 83.7 & 39.2 & 22.2 & 28.5 & 36.0 & 17.0 & 84.2 & 35.9 & 85.8 & 59.1 & 35.5 & 85.2 & 31.1 & 39.3 & 21.1 & 26.7 & 27.5 & 46.6 \\
CCM\cite{Li2020} &  93.5 & 57.6 & 84.6 & 39.3 & 24.1 & 25.2 & 35.0 & 17.3 & 85.0 & 40.6 & 86.5 & 58.7 & 28.7 & 85.8 & \textbf{49.0} & 56.4 & 5.4 & 31.9 & 43.2 & 49.9 \\
FADA\cite{wang2020a} & 91.0 & 50.6 & 86.0 & 43.4 & 29.8 & 36.8 & 43.4 & 25.0 & 86.8 & 38.3 & 87.4 & 64.0 & 38.0 & 85.2 & 31.6 & 46.1 & 6.5 & 25.4 & 37.1 & 50.1\\
CAG\cite{Zhang2019} & 90.4 & 51.6 & 83.8 & 34.2 & 27.8 & 38.4 & 25.3 & 48.4 & 85.4 & 38.2 & 78.1 & 58.6 & 34.6 & 84.7 & 21.9 & 42.7 & \textbf{41.1} & 29.3 & 37.2 & 50.2\\
PIT\cite{Lv2020} & 87.5 & 43.4 & 78.8 & 31.2 & 30.2 & 36.3 &  39.9 &  42.0 & 79.2 &  37.1 & 79.3 & 65.4 & 37.5 & 83.2 & 46.0 & 45.6 & 25.7 & 23.5 & 49.9 & 50.6 \\
DACS\cite{Tranheden2020} & 89.9 & 39.7 & 87.9 & 30.7 & 39.5 & 38.5 & 46.4 & 52.8 & 88.0 & 44.0 & 88.8 & 67.2 & 35.8 & 84.5 & 45.7 & 50.2 & 0.00 & 27.3 & 34.0 & 52.1 \\
MRnet\cite{Zheng2020a} & 90.4 & 31.2 & 85.1 & 36.9 & 25.6 & 37.5 & 48.8 & 48.5 & 85.3 & 34.8 & 81.1 & 64.4 & 36.8 & 86.3 & 34.9 & 52.2 & 1.7 & 29.0 & 44.6 & 50.3 \\
IAST\cite{Mei2020} & \textbf{94.1} & \textbf{58.8} & 85.4 & 39.7 & 29.2 & 25.1 & 43.1 & 34.2 & 84.8 & 34.6 & \textbf{88.7} & 62.7 & 30.3 & 87.6 & 42.3 & 50.3 & 24.7 & 35.2 & 40.2 & 52.2 \\
Meta\cite{Guo2021} & 92.8 & 58.1 & 86.2 & 39.7 & 33.1 & 36.3 & 42.0 & 38.6 & 85.5 & 37.8 & 87.6 & 62.8 & 31.7 & 84.8 & 35.7 & 50.3 & 2.0 & 36.8 & 48.0 & 52.1 \\

Pix\cite{MelasKyriazi2021} & 91.6 & 51.2 & 84.7 & 37.3 & 29.1 & 24.6 & 31.3 & 37.2 & 86.5 & 44.3 & 85.3 & 62.8 & 22.6 & 87.6 & 38.9 & 52.3 & 0.65 & 37.2 & 50.0 & 50.3 \\
  \hline
Ours & 93.2 & 53.1 & \textbf{86.8} & 40.6 & 35.7 & 37.2 & 42.4 & 48.7 & 86.1 & 35.5 & 84.7 & 67.6 & 34.7 & 88.3 & 47.6 & 46.7 & 3.8 & 39.8 & 54.8 & 54.1 \\
    \hline
\end{tabular}
}
\end{table*}

\begin{table*} 
 \centering
  \caption{Comparison between EPS-UDA and other state-of-the-art methods  on SYNTHIA $\to$ Cityscapes. mIoU* represents 13 classes, and mIoU represents 16 classes. }
  \label{tab:synthia}   
    \resizebox{\textwidth}{!}{
  \setlength{\tabcolsep}{0.009\textwidth}
  \begin{tabular}{l|ccc ccc ccc ccc ccc c|c |c}

    \hline
    method & road & S.W. & build & wall &  fence & pole & light & sign & Veg.& sky & P.R. & rider & car & bus &  motor & bike & mIoU* & mIoU\\
    \hline
Source\cite{Tsai2018}& 55.6 & 23.8 & 74.6 & - & - & - & 6.1 & 12.1 & 74.8 & 79.0 & 55.3 & 19.1 & 39.6 & 23.3 & 13.7 & 25.0 & 38.7 & -\\
Adapt\cite{Tsai2018} &84.3 & 42.7 & 77.5 & - & - & - & 4.7 & 7.0 & 77.9 & 82.5 & 54.3 & 21.0 & 72.3 & 32.2 & 18.9 & 32.3 & 46.7 & - \\
FADA\cite{wang2020a} & 84.5 & 40.1 & 83.1 & 4.8 & 0.0 & 34.3 & 20.1 & 27.2 & 84.8 & 84.0 & 53.5 & 22.6 & 85.4 & 43.7 & 26.8 & 27.8  & 52.5 & 45.2\\
SIM\cite{wang2020} &83.0 & 44.0 & 80.3 & - & - & - & 17.1 & 15.8 & 80.5 & 81.8 & 59.9 & 33.1 & 70.2 & 37.3 & 28.5 & 45.8 & 52.1 & - \\
MRnet\cite{Zheng2020a}& 87.6 & 41.9 & 83.1 & 14.7 & 1.7 & 36.2 & 31.3 & 19.9 & 81.6 & 80.6 & 63.0 & 21.8 & 86.2 & 40.7 & 23.6 & 53.1 & 54.9 & 47.9\\
IAST\cite{Mei2020} & 81.9 & 41.5 & 83.3 & 17.7 & 4.6 & 32.3 & 30.9 & 28.8 & 83.4 & 85.0 & 65.5 & 30.8 & 86.5 & 38.2 & 33.1 & 52.7 & 57.0  & 49.8\\
CAG\cite{Zhang2019} & 84.7 & 40.8 & 81.7 & 7.8 & 0.0 & 35.1 & 13.3 & 22.7 & 84.5 & 77.6 & 64.2 & 27.8 & 80.9 & 19.7 & 22.7 & 48.3 & 51.5 & 44.5  \\
PIT\cite{Lv2020} & 83.1 & 27.6 & 81.5 & 8.9 & 0.3 & 21.8 & 26.4 & 33.8 & 76.4 & 78.8 & 64.2 & 27.6 & 79.6 & 31.2 & 31.0 & 31.3 & 51.8 & 44.0  \\
DACS\cite{Tranheden2020} & 80.6 & 25.1 & 81.9 & 21.5 & 2.9 & 37.2 & 22.8 & 24.0 & 83.7 & 90.8 & 67.6 & 38.3 & 82.9 & 38.9 & 28.5 & 47.6 & 54.8 & 48.3 \\
Meta\cite{Guo2021} &92.6 & 52.7 & 81.3 & 8.9 & 2.4 & 28.1 & 13.0 & 7.3 & 83.5 & 85.0 & 60.1 & 19.7 & 84.8 & 37.2 & 21.5 & 43.9 & 52.5 & 45.1 \\
Pix\cite{MelasKyriazi2021} & 92.5 & 54.6 & 79.8 & 4.8 & 0.1 & 24.1 & 22.8 & 17.8 & 79.4 & 76.5 & 60.8 & 24.7 & 85.7 & 33.5 & 26.4 & 54.4 & 54.5 & 46.1 \\
  \hline
Ours & 89.2 & 49.5 & 81.8 & 9.3 & 1.7 & 37.4 & 33.8 & 29.0 & 83.5 & 85.4 & 64.5 & 30.6 & 84.4 & 47.8 & 20.0 & 53.0 & 57.9 & 50.1 \\

\hline
\end{tabular}
}
\end{table*}
\newcommand{\pic}[1]{\includegraphics[width=0.19\textwidth]{#1}}
\begin{figure*}
 \setlength\tabcolsep{1pt}
    \centering
    \begin{tabular}{*{5}{c}}
        \pic{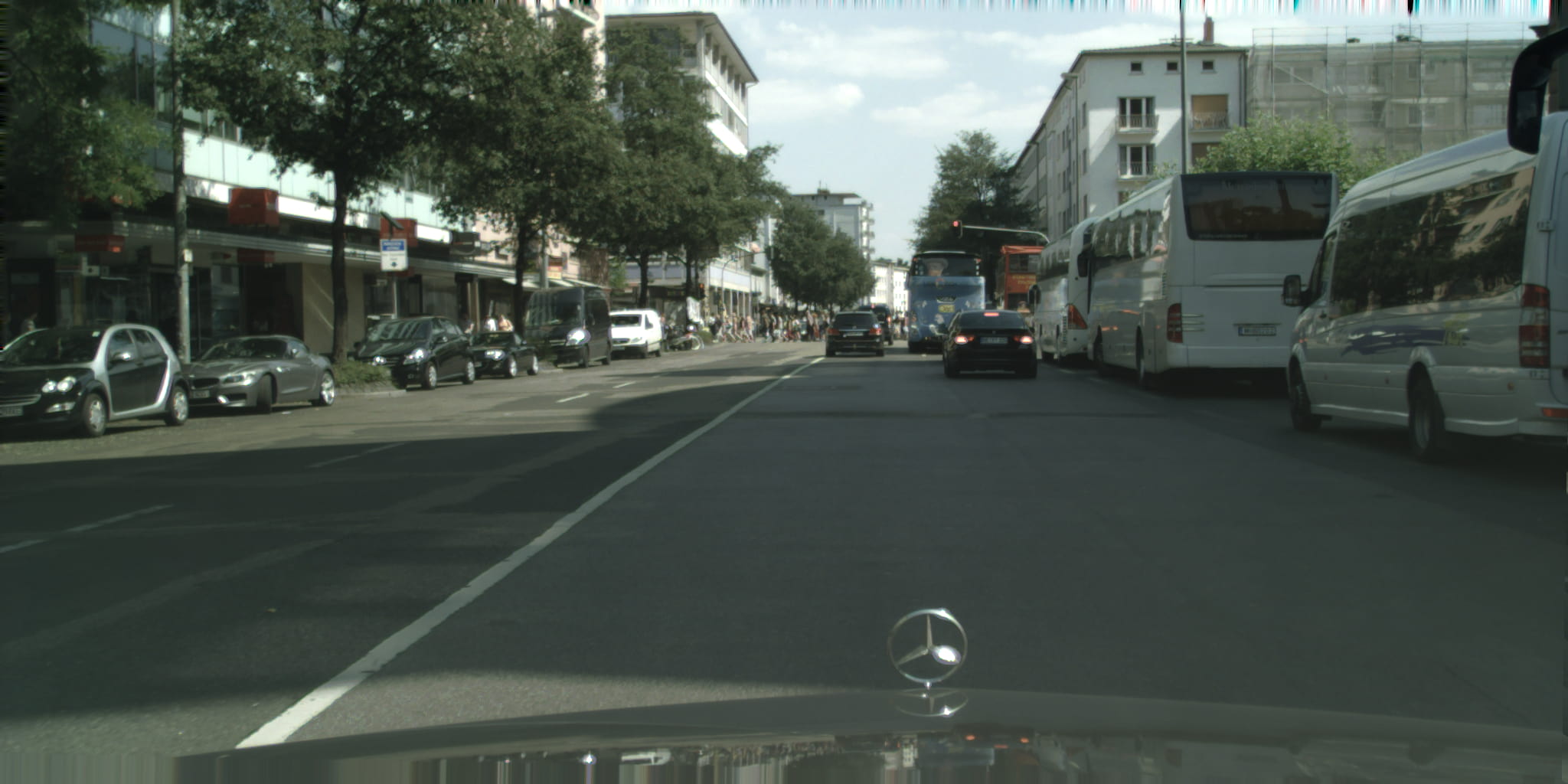} & \pic{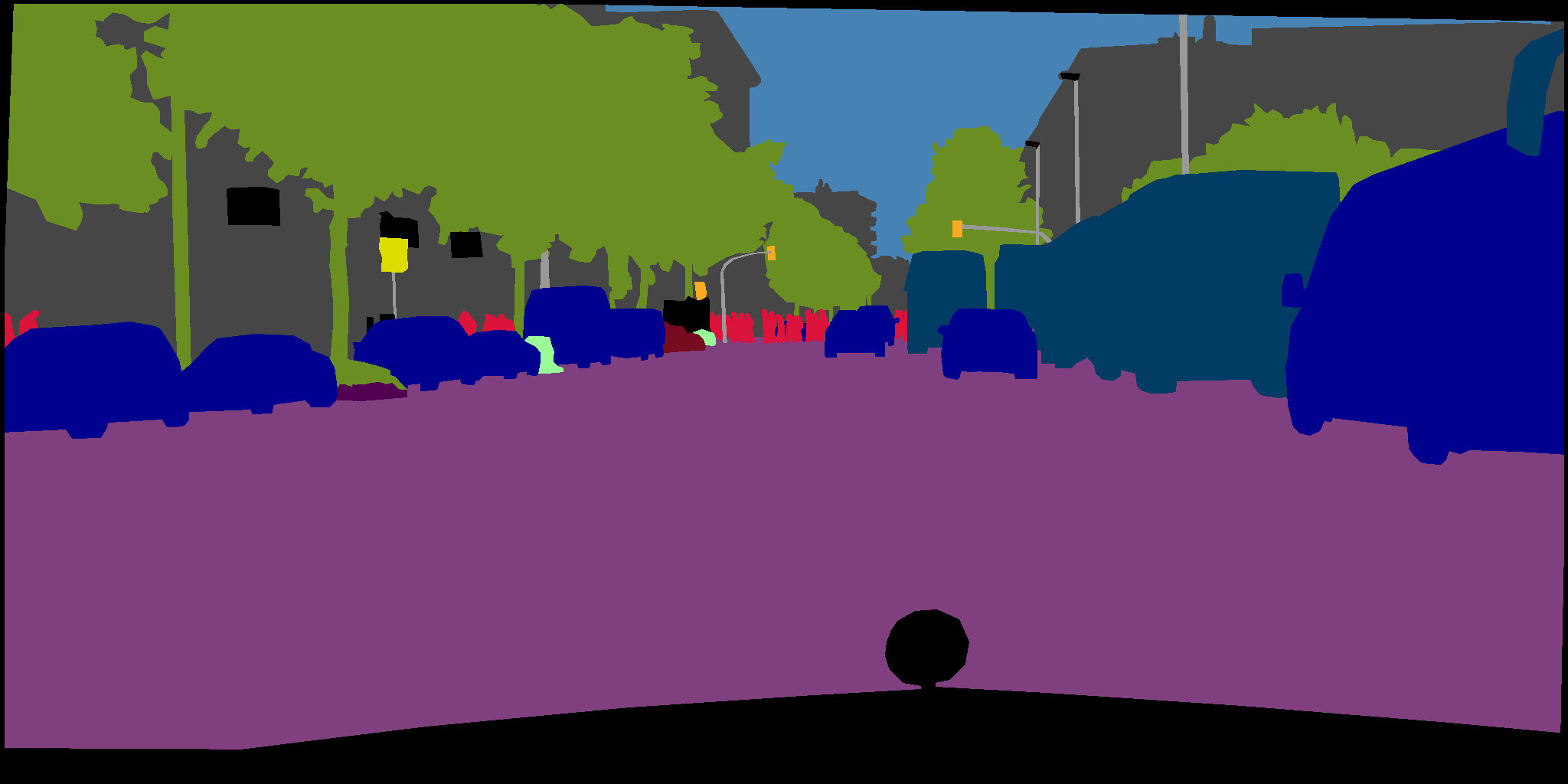}  & \pic{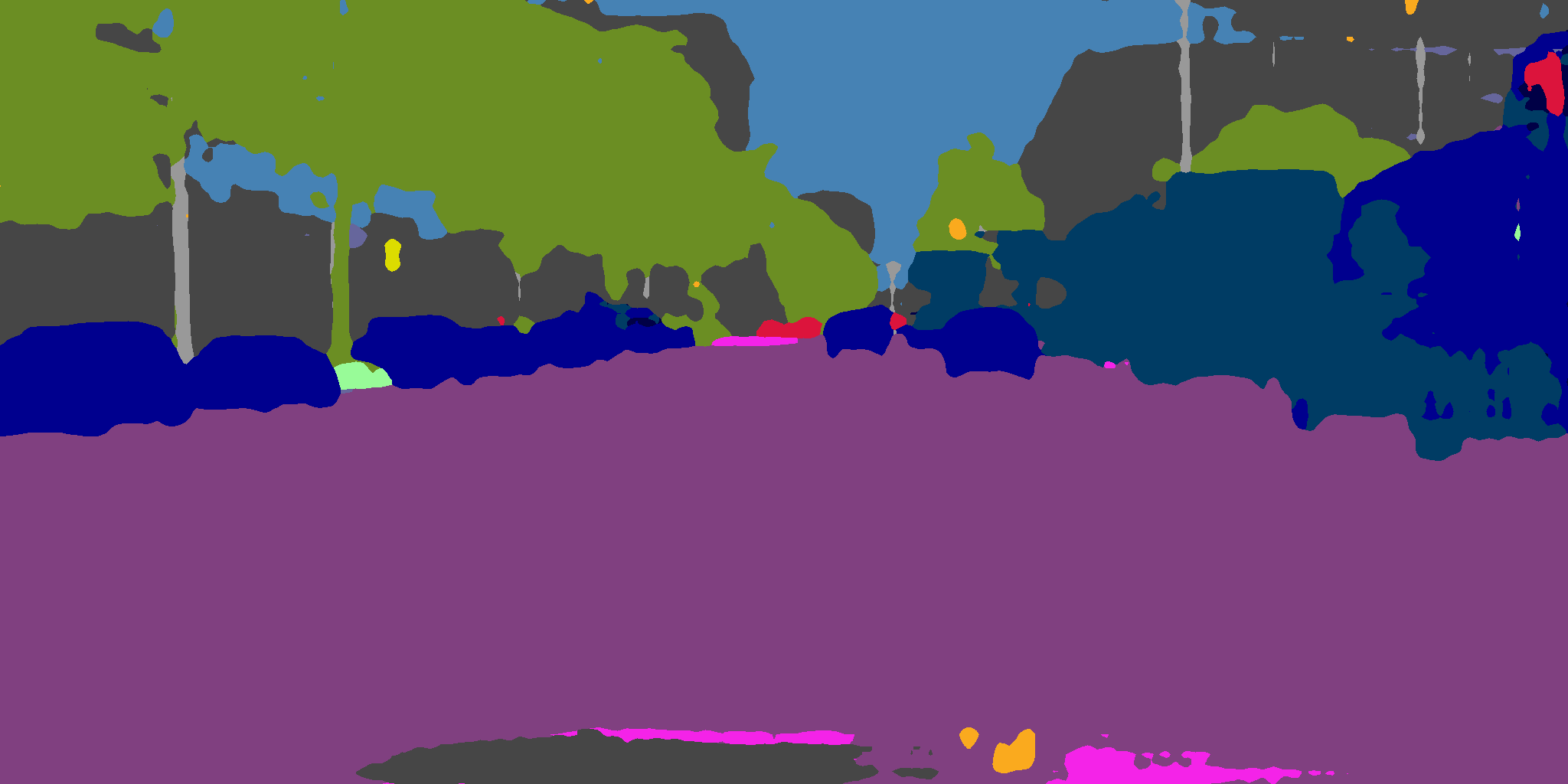}  & \pic{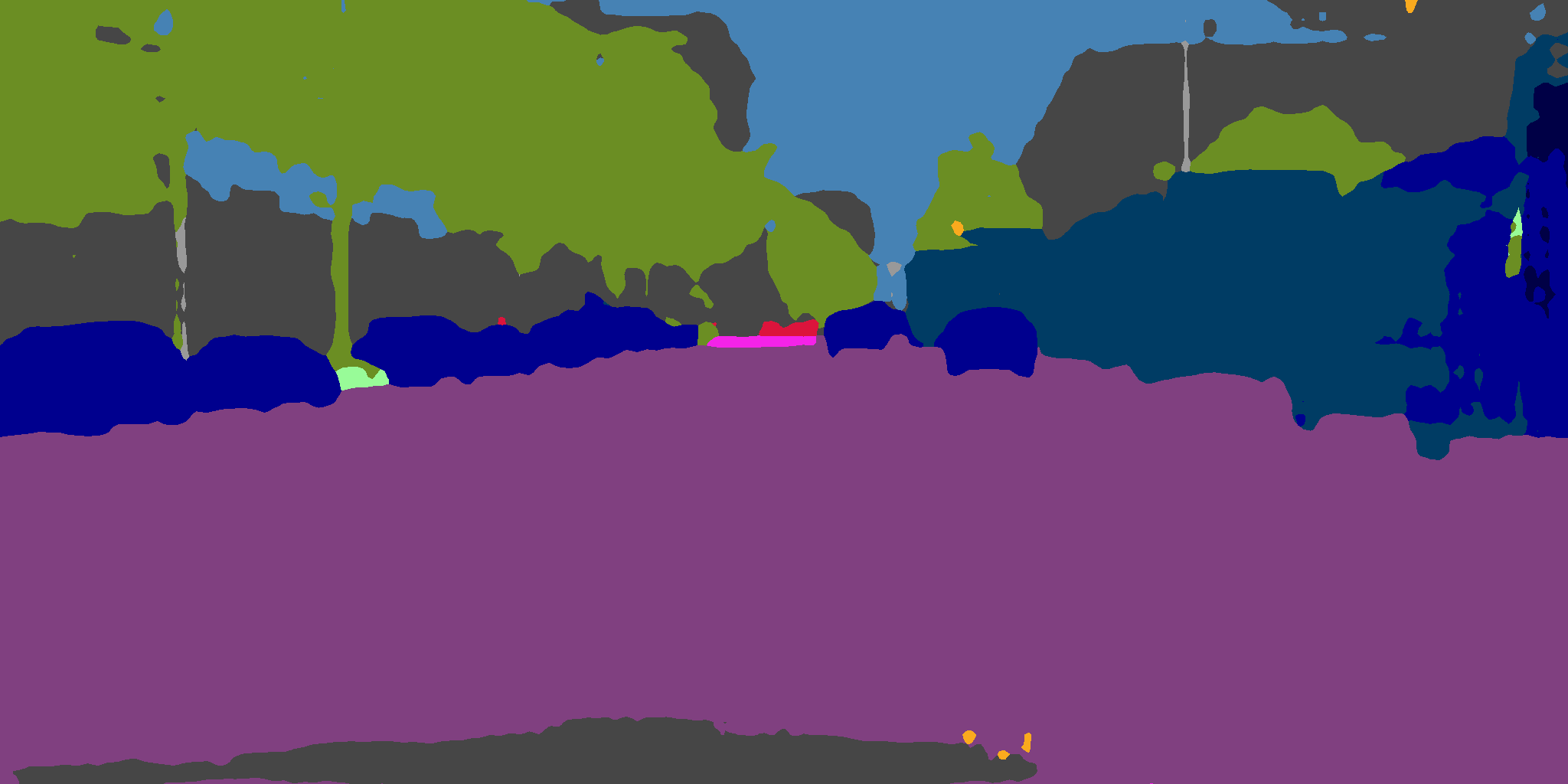}  & \pic{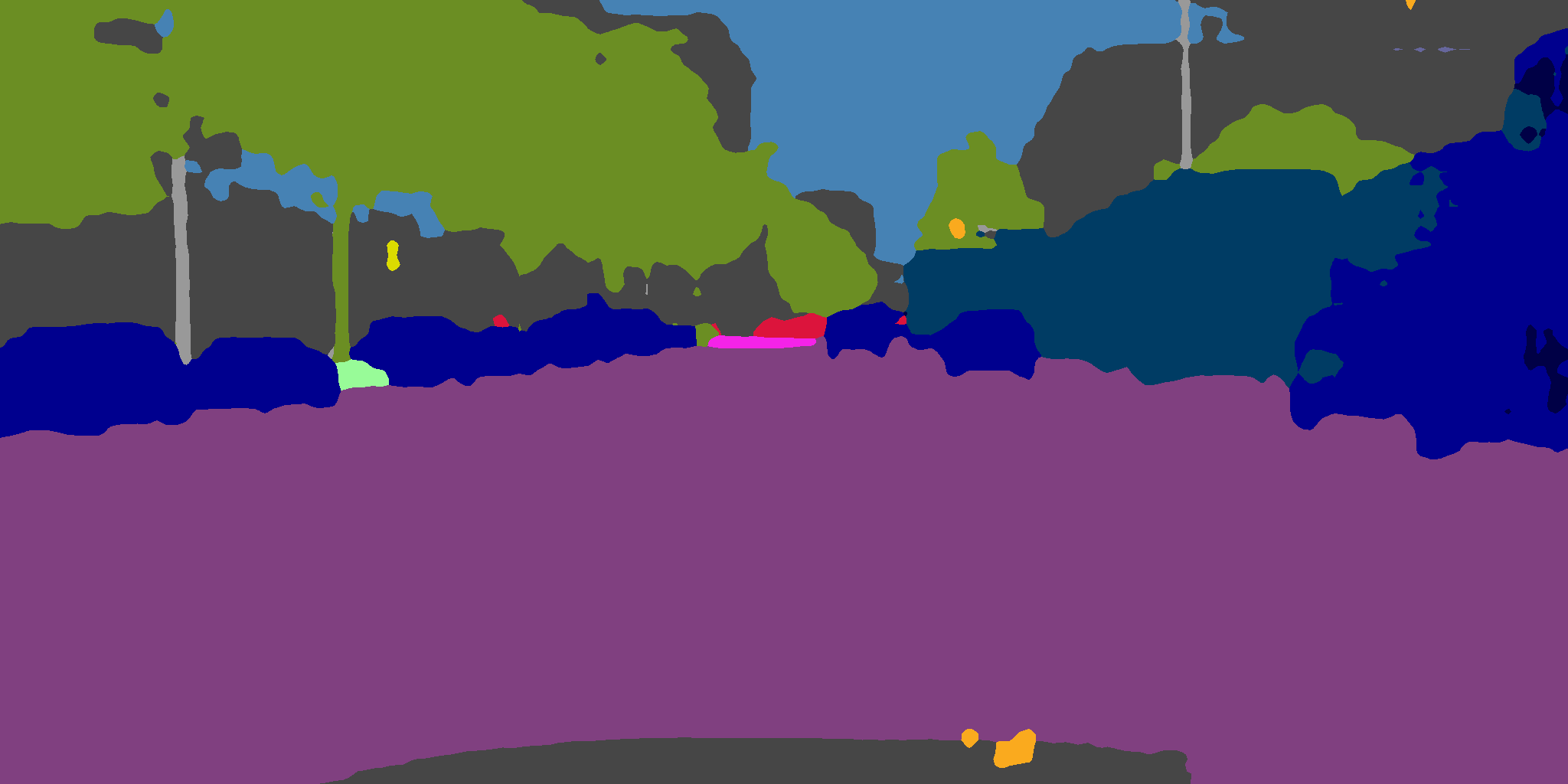} \\
        \pic{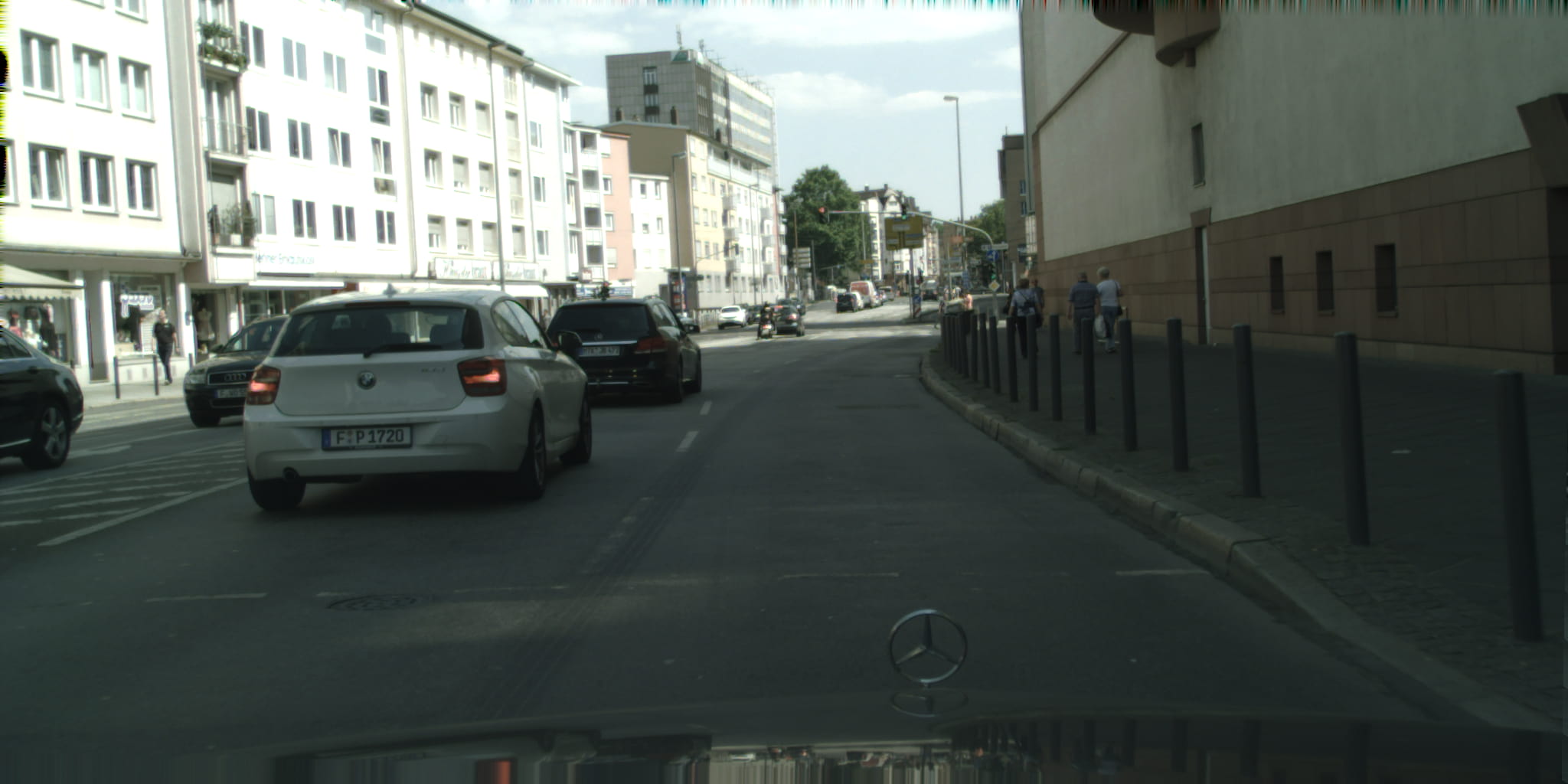} & \pic{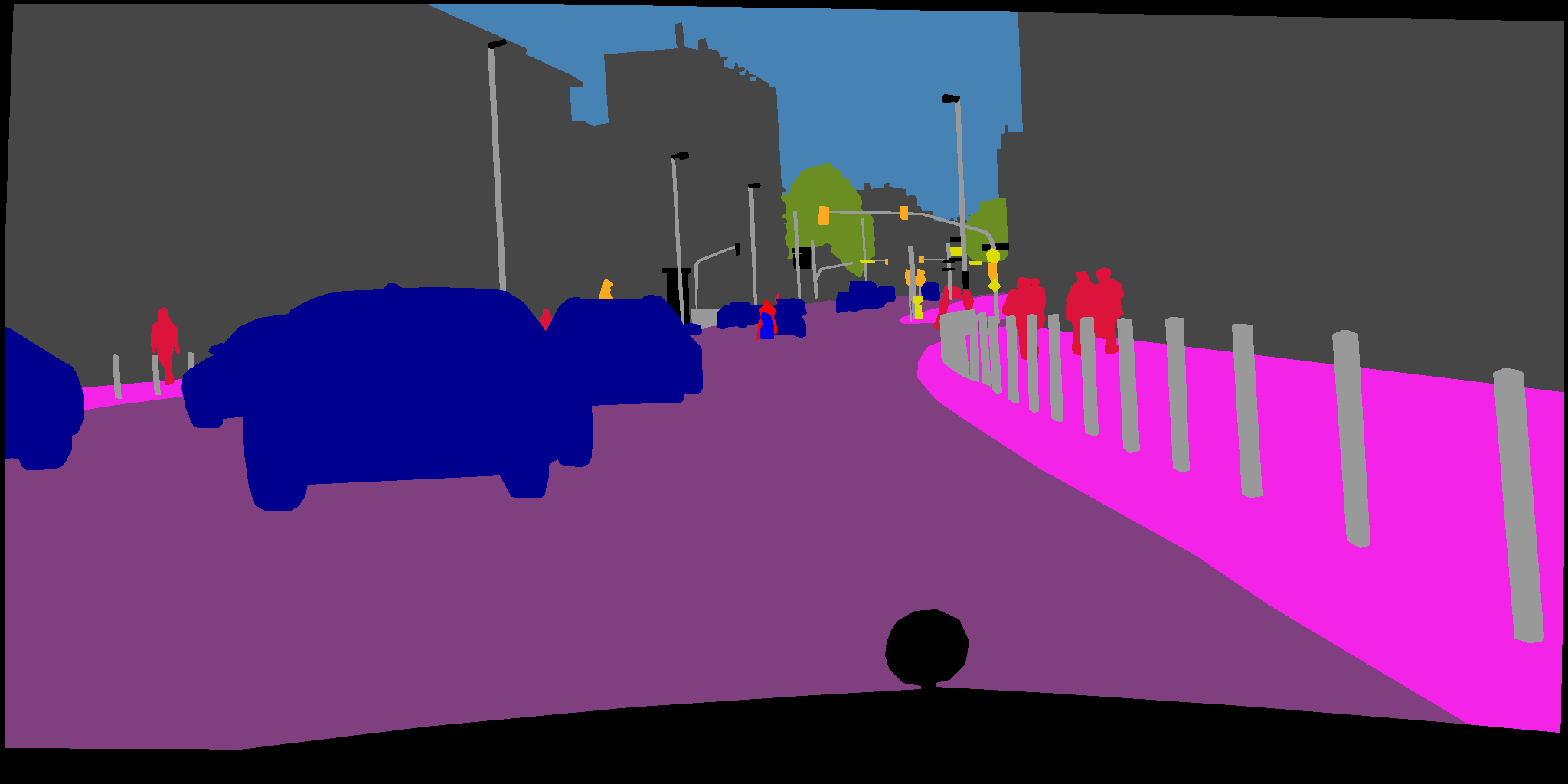}  & \pic{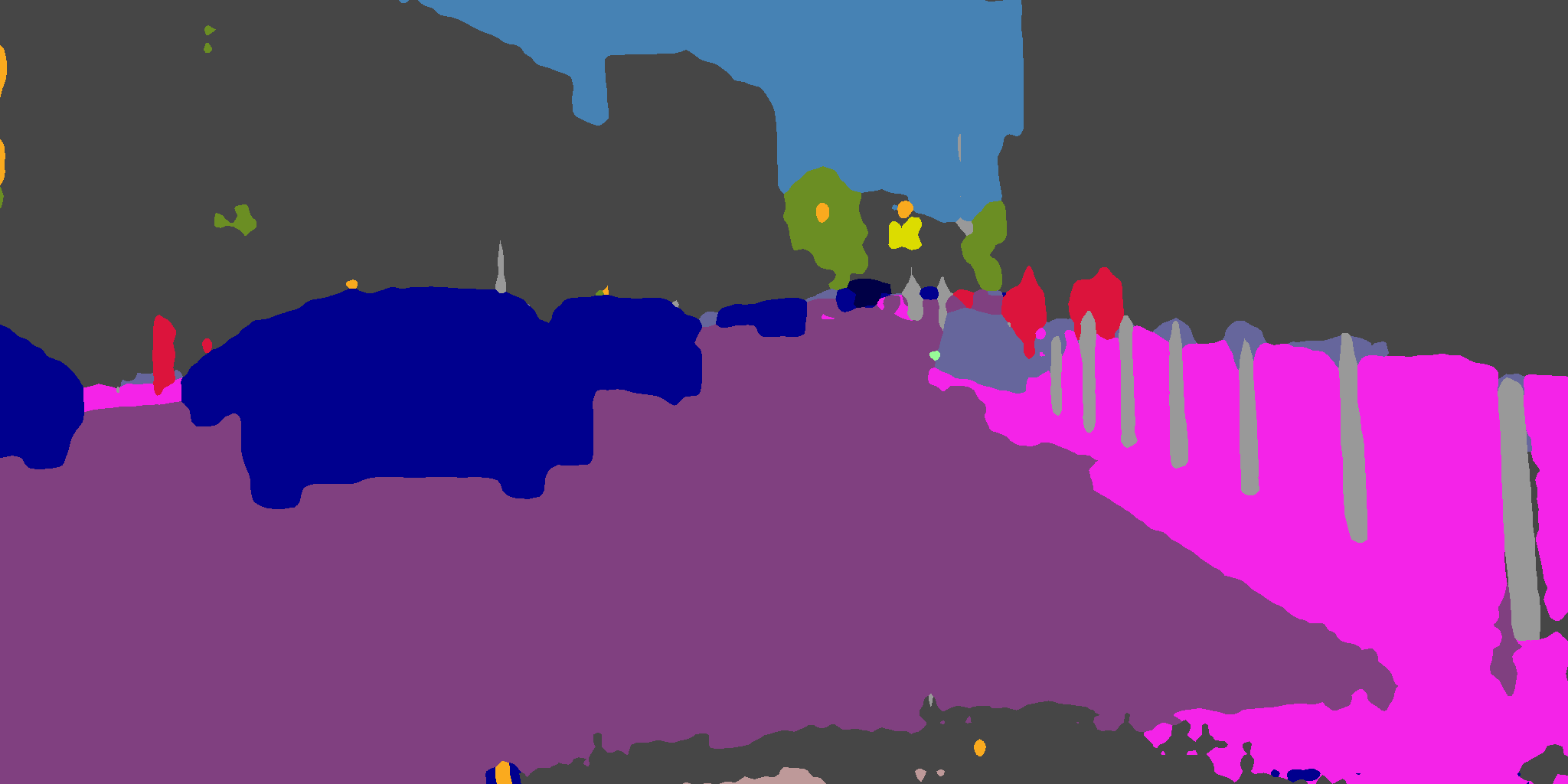}  & \pic{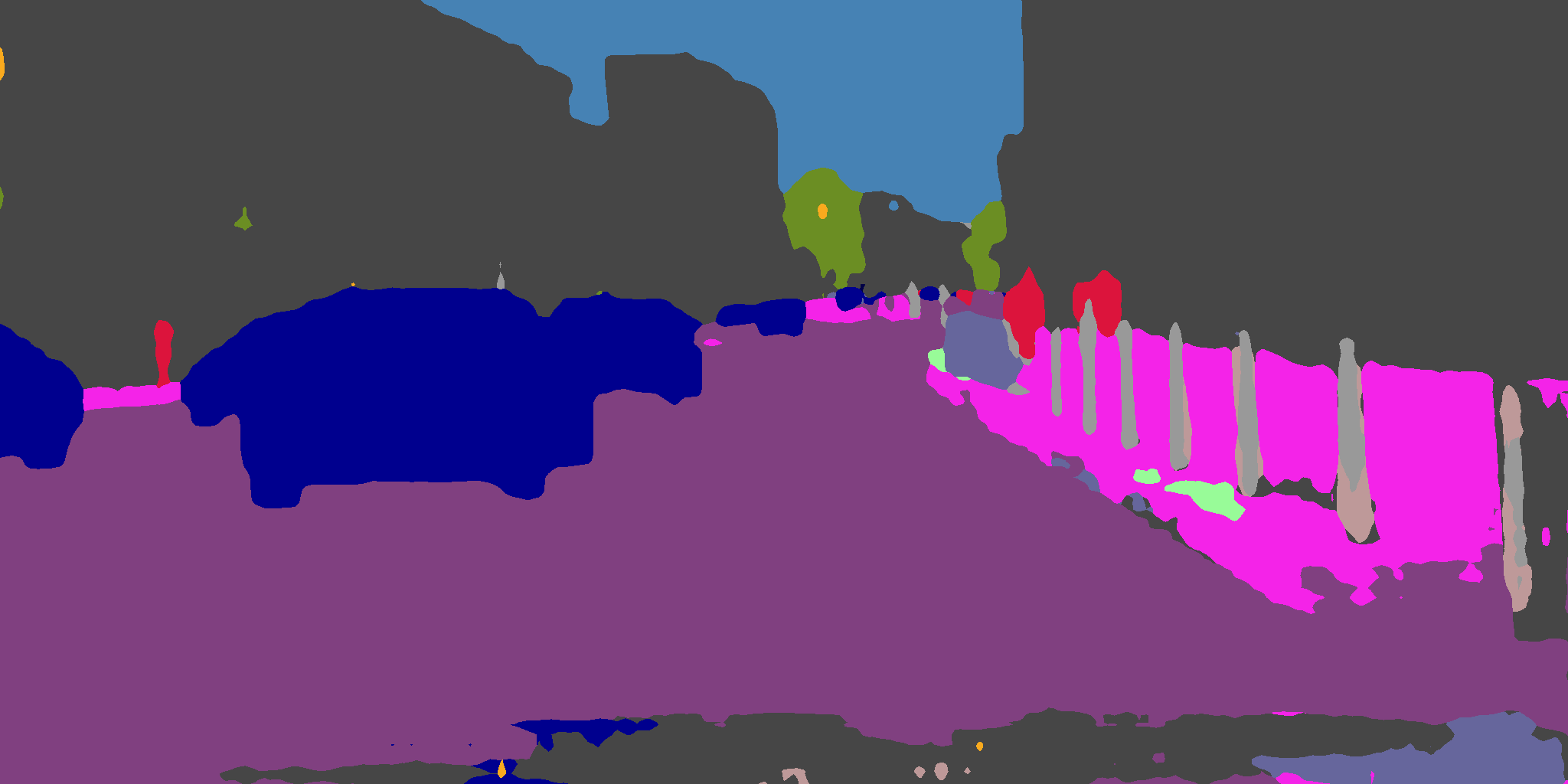}  & \pic{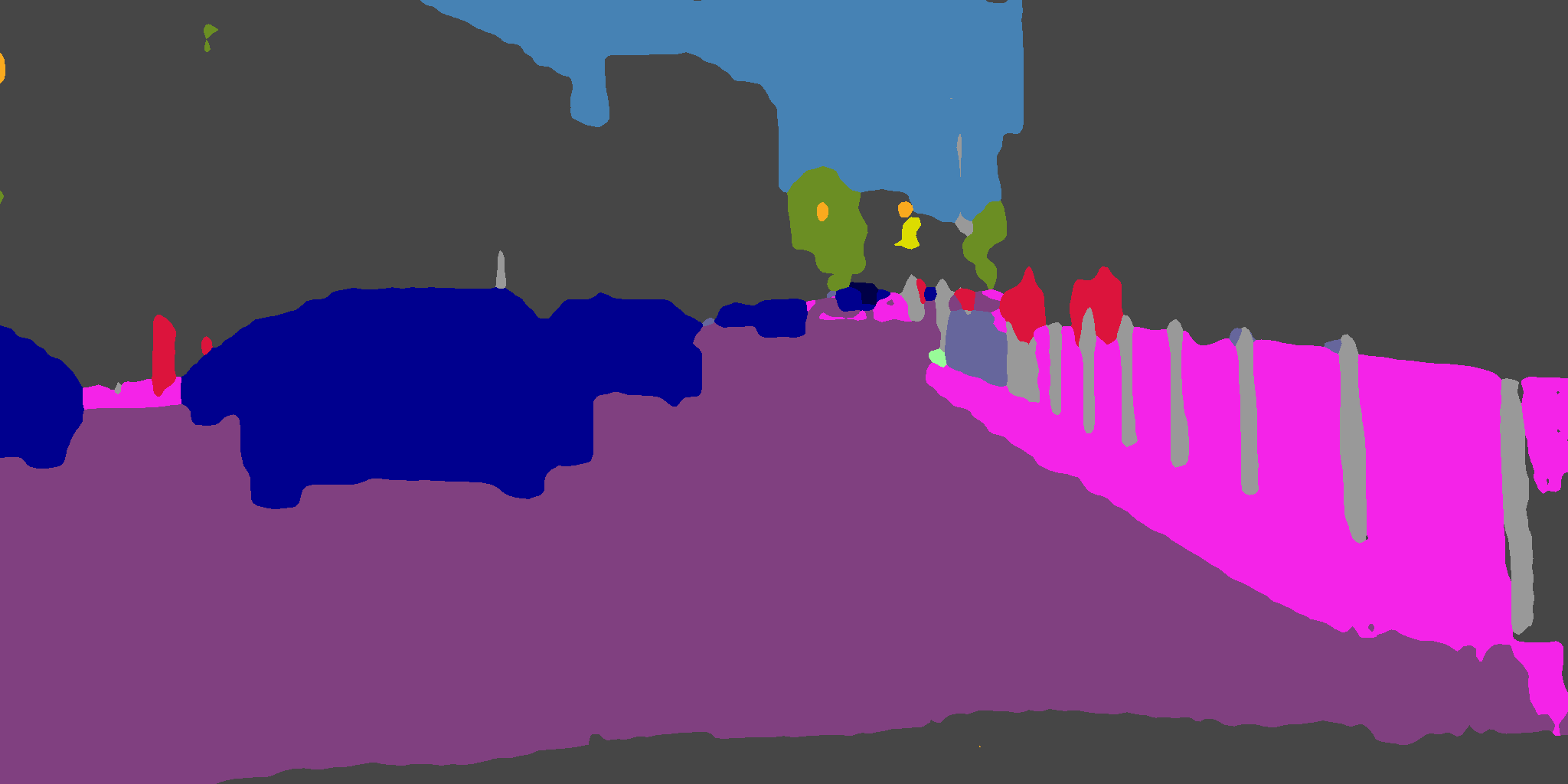} \\
        \pic{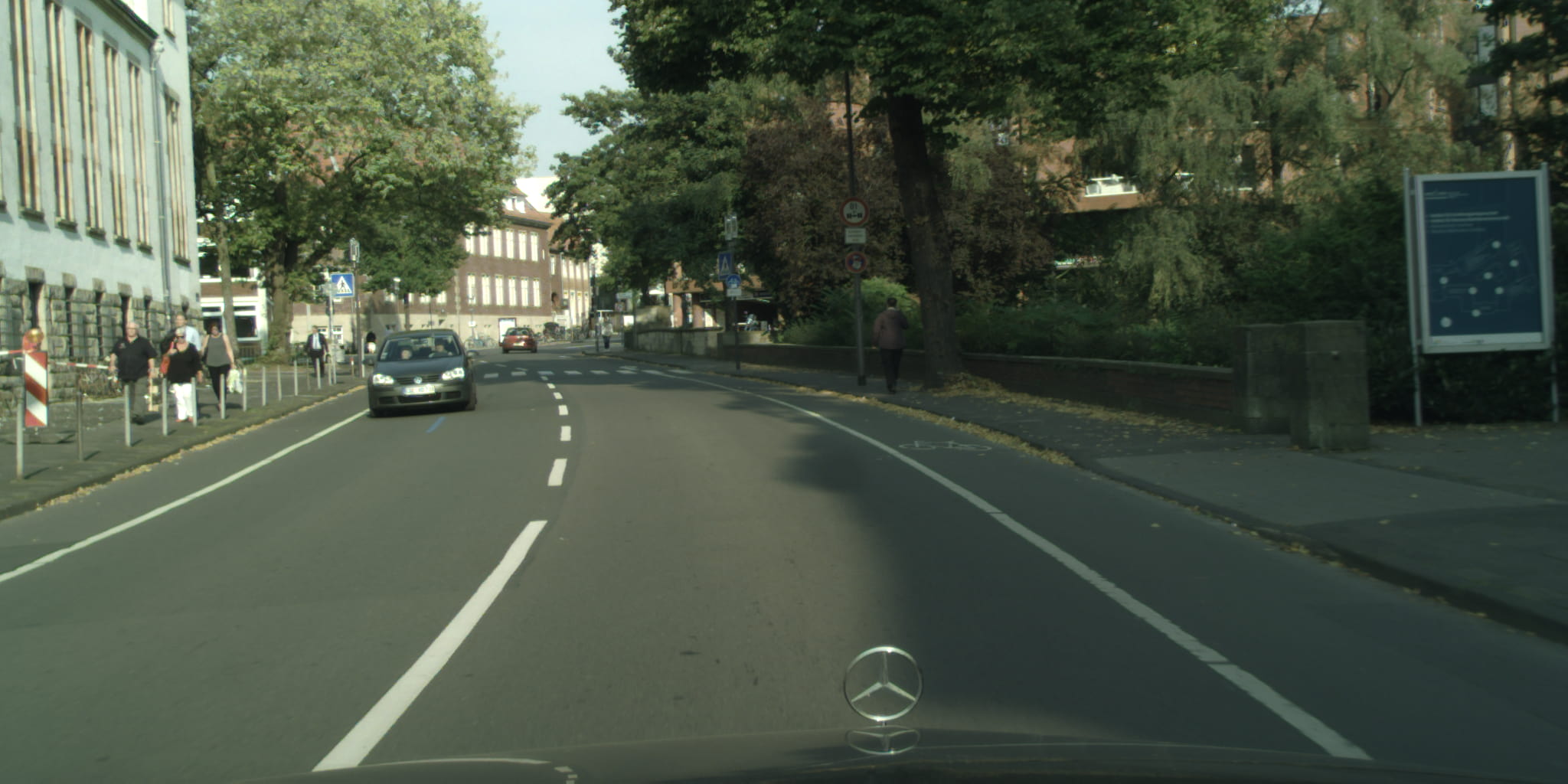} & \pic{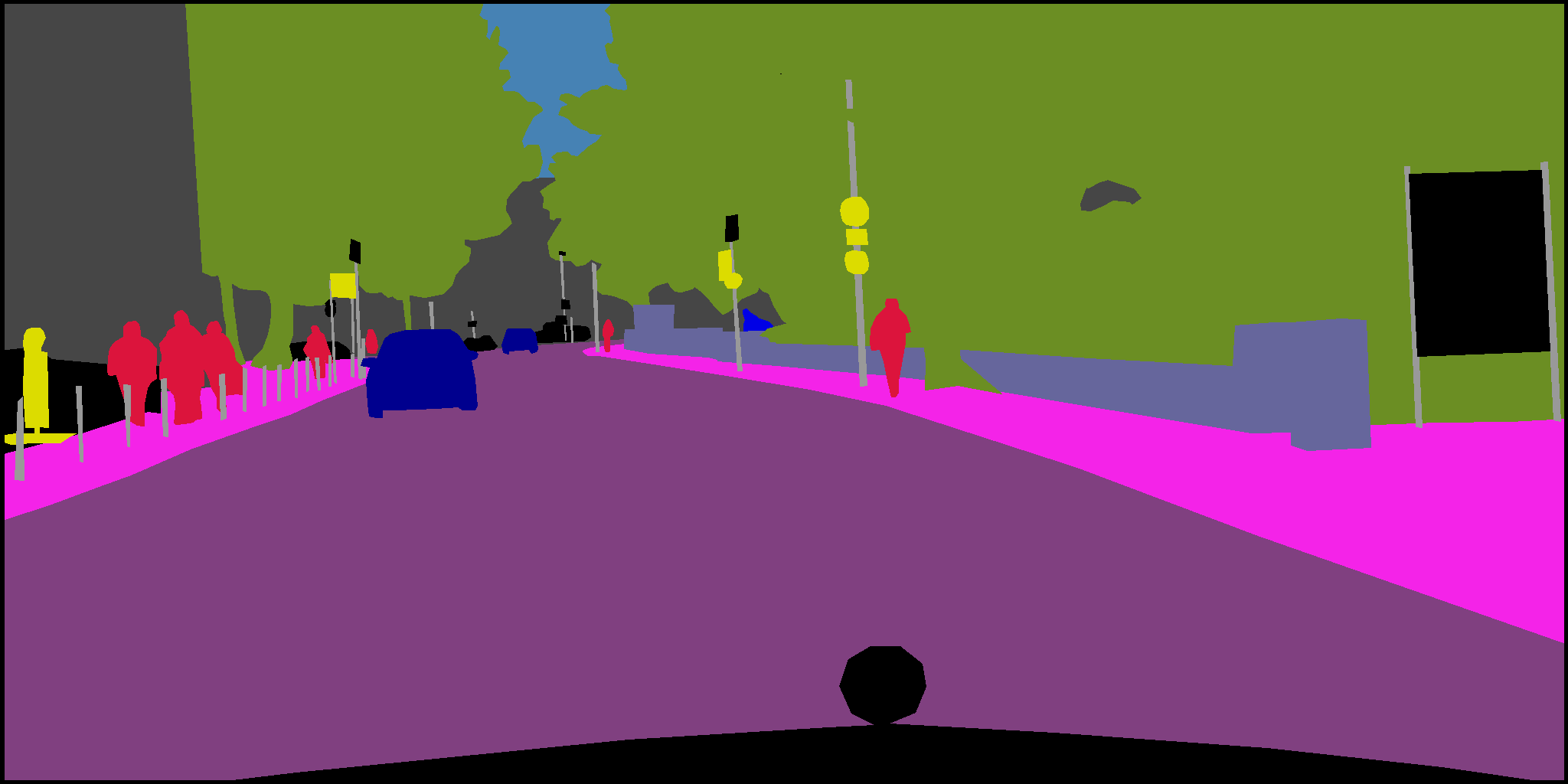}  & \pic{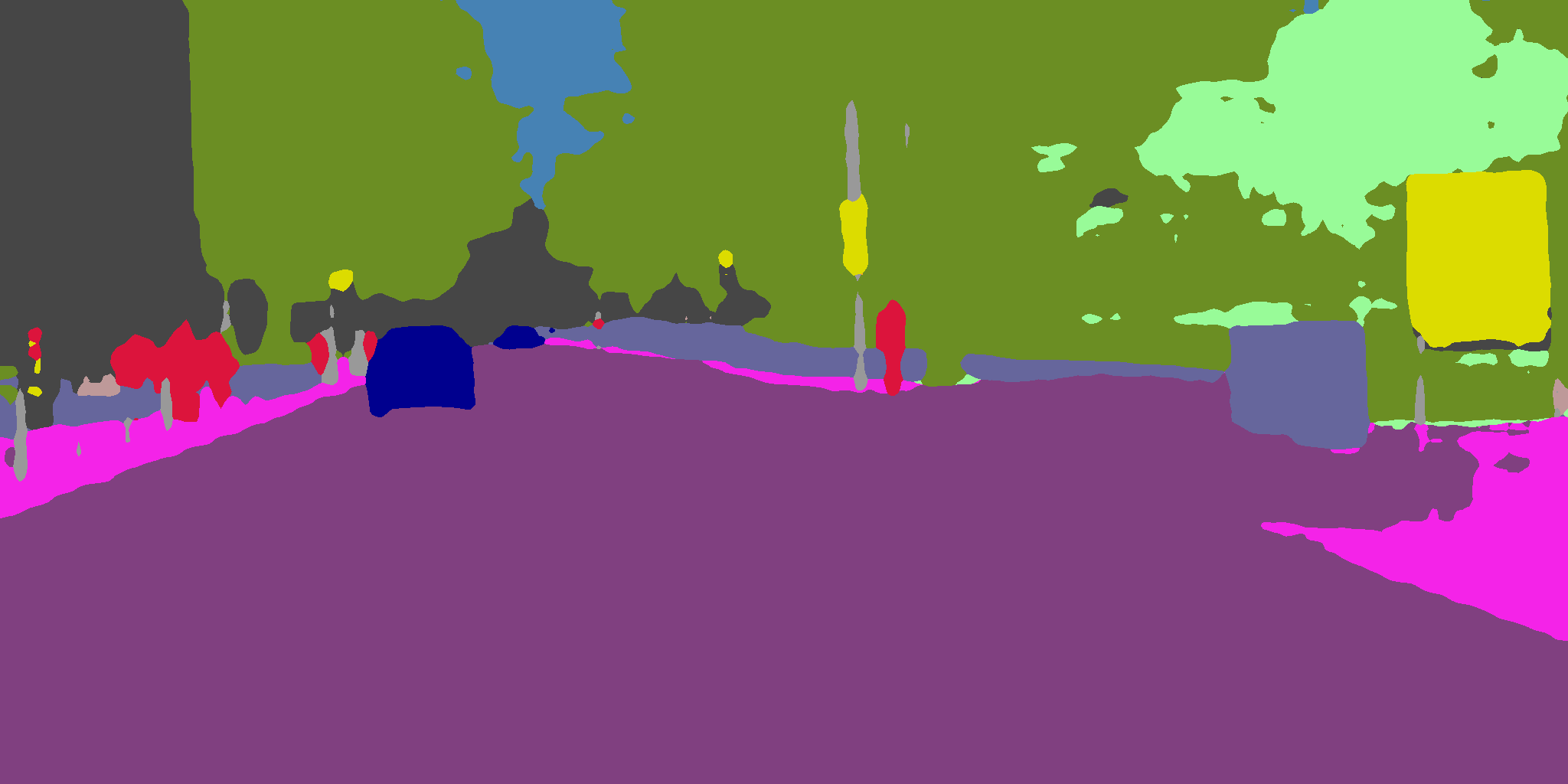}  & \pic{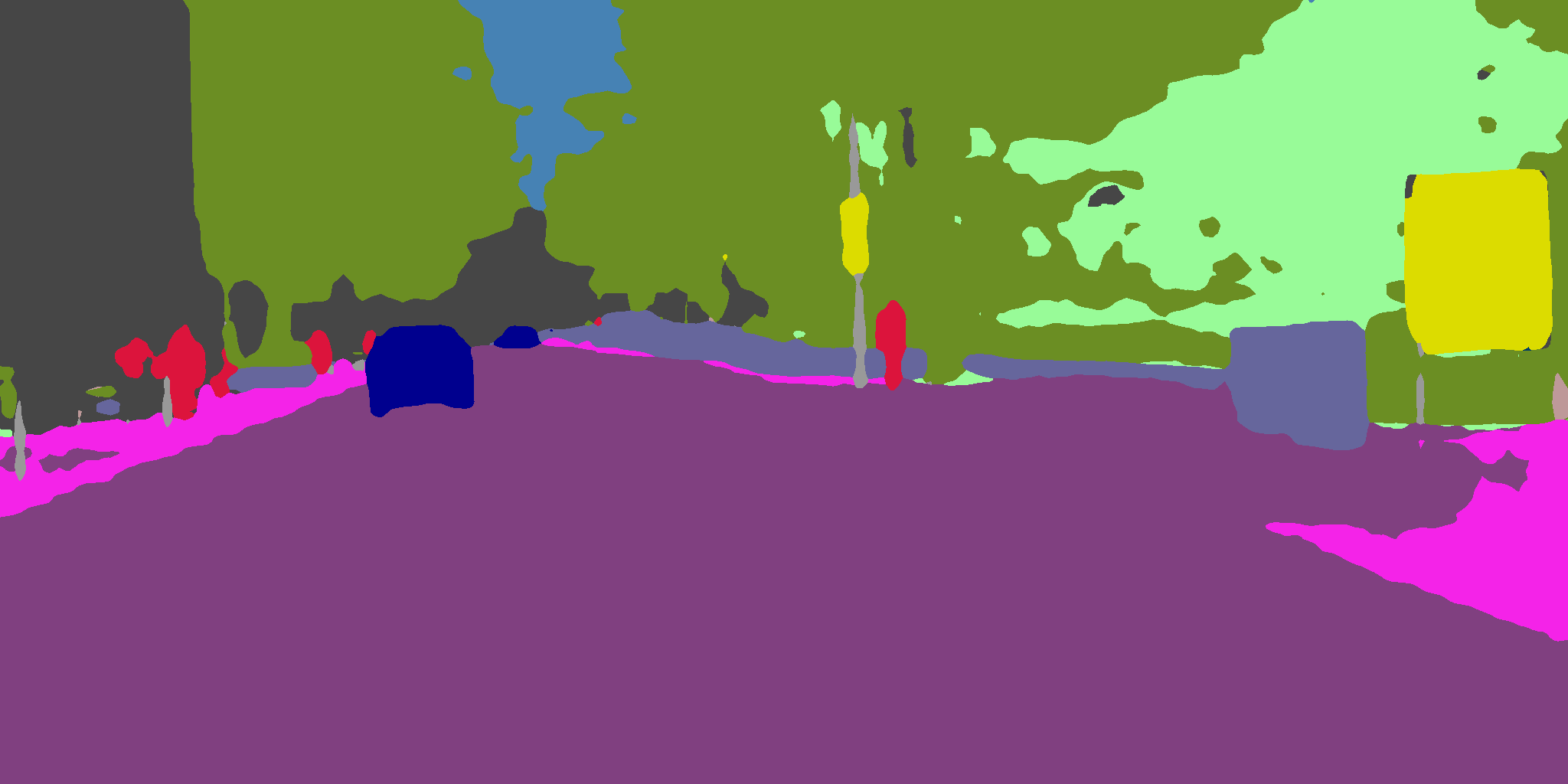}  & \pic{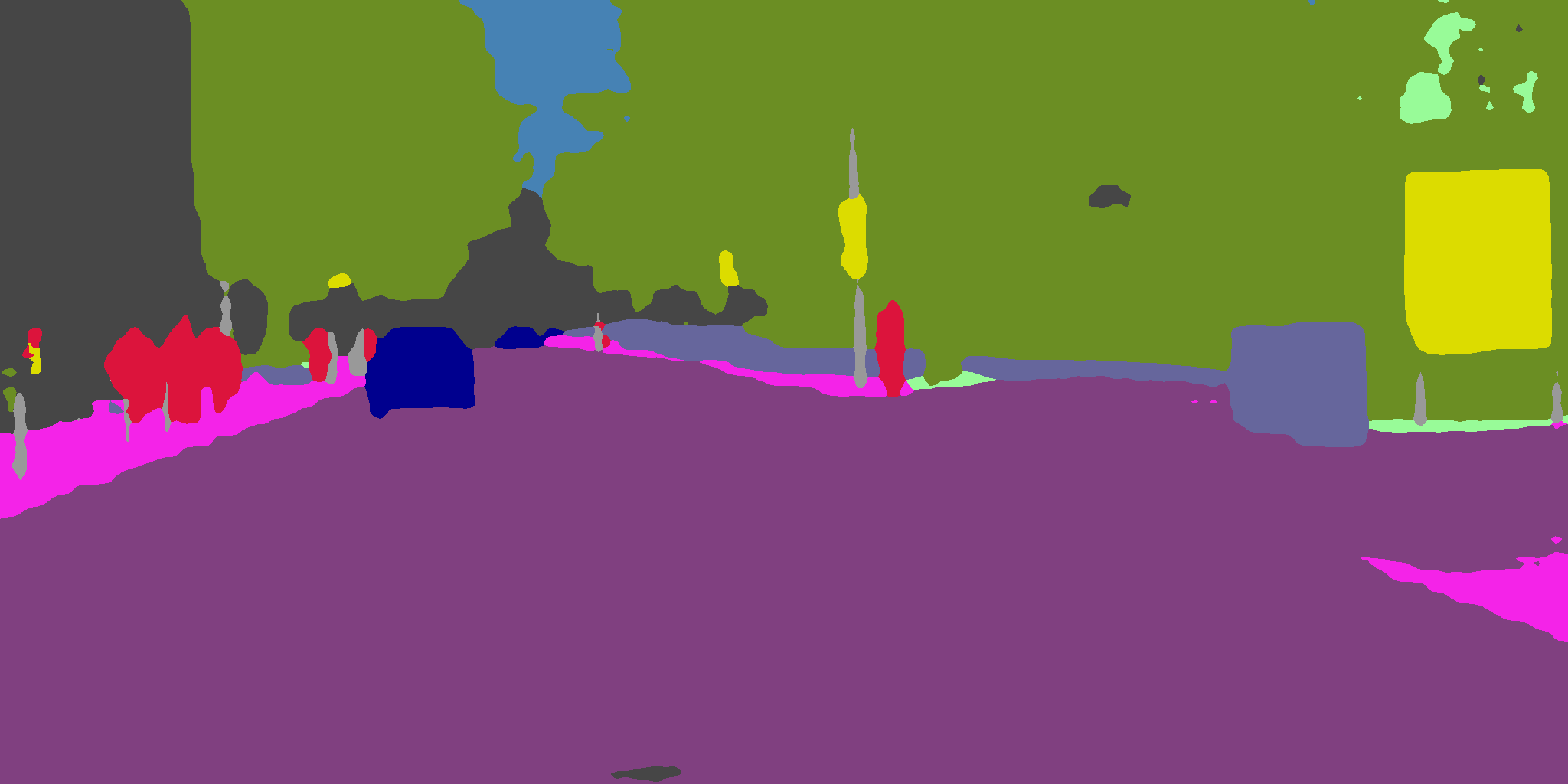} \\
        \pic{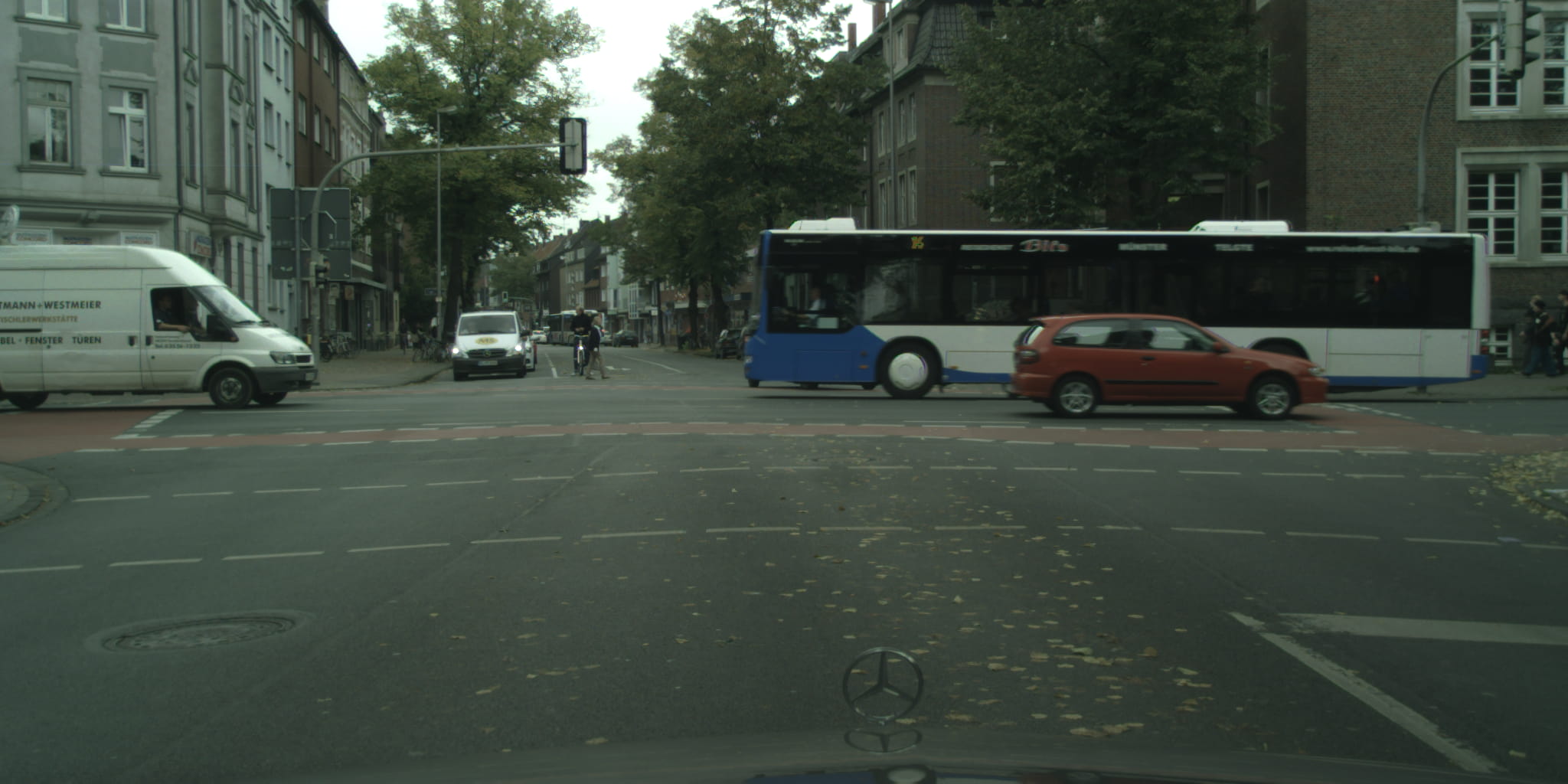} & \pic{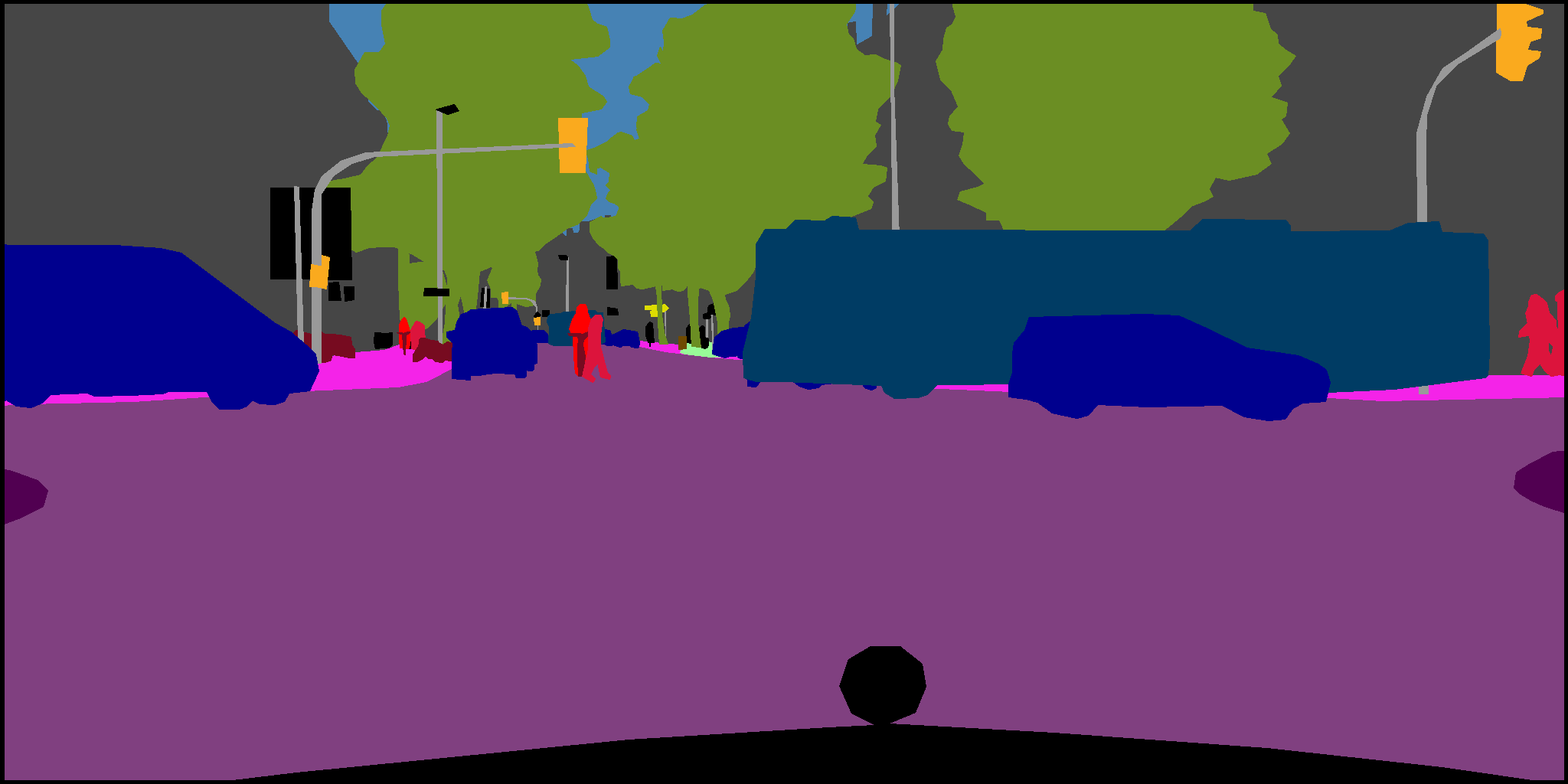}  & \pic{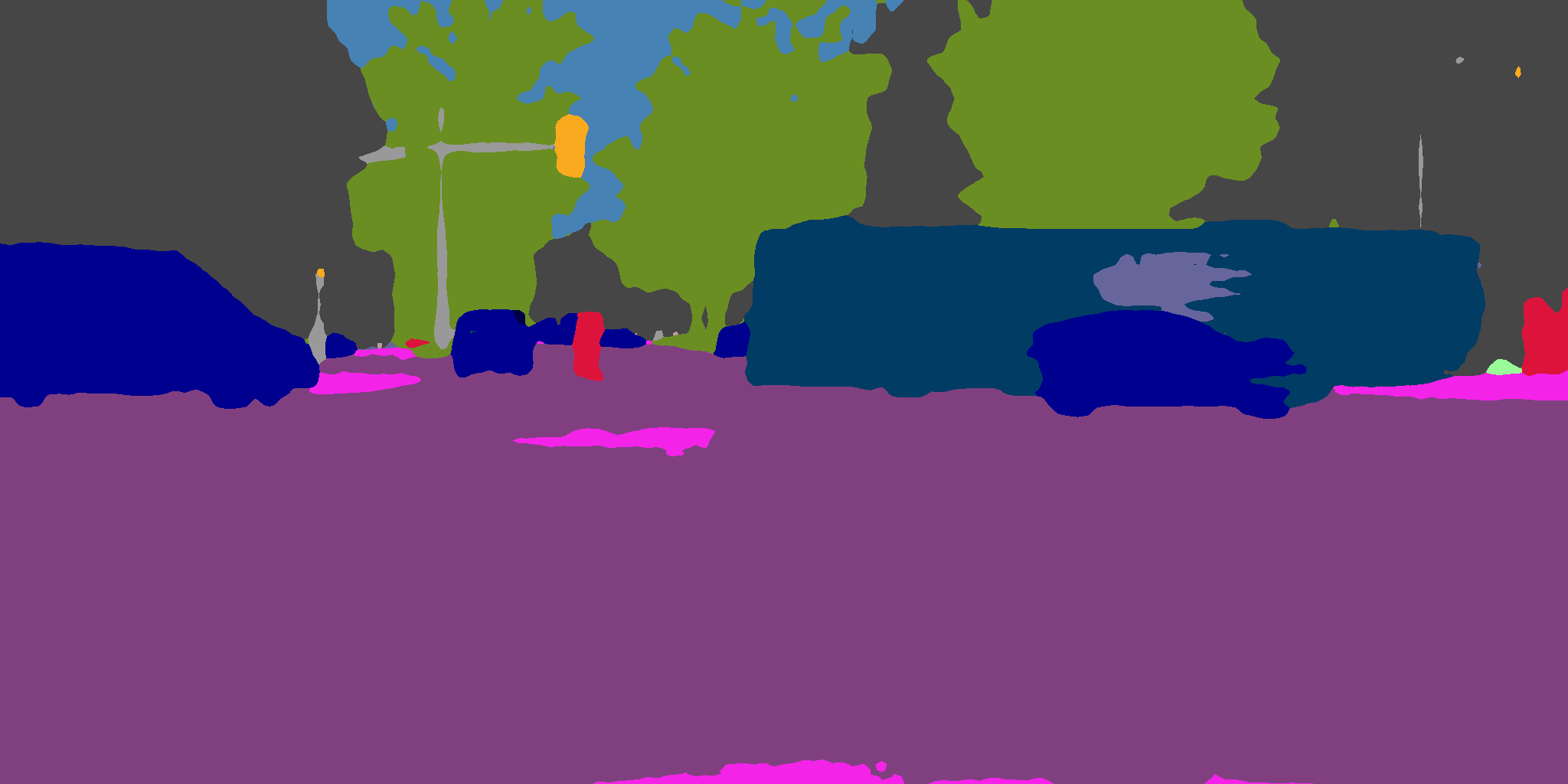}  & \pic{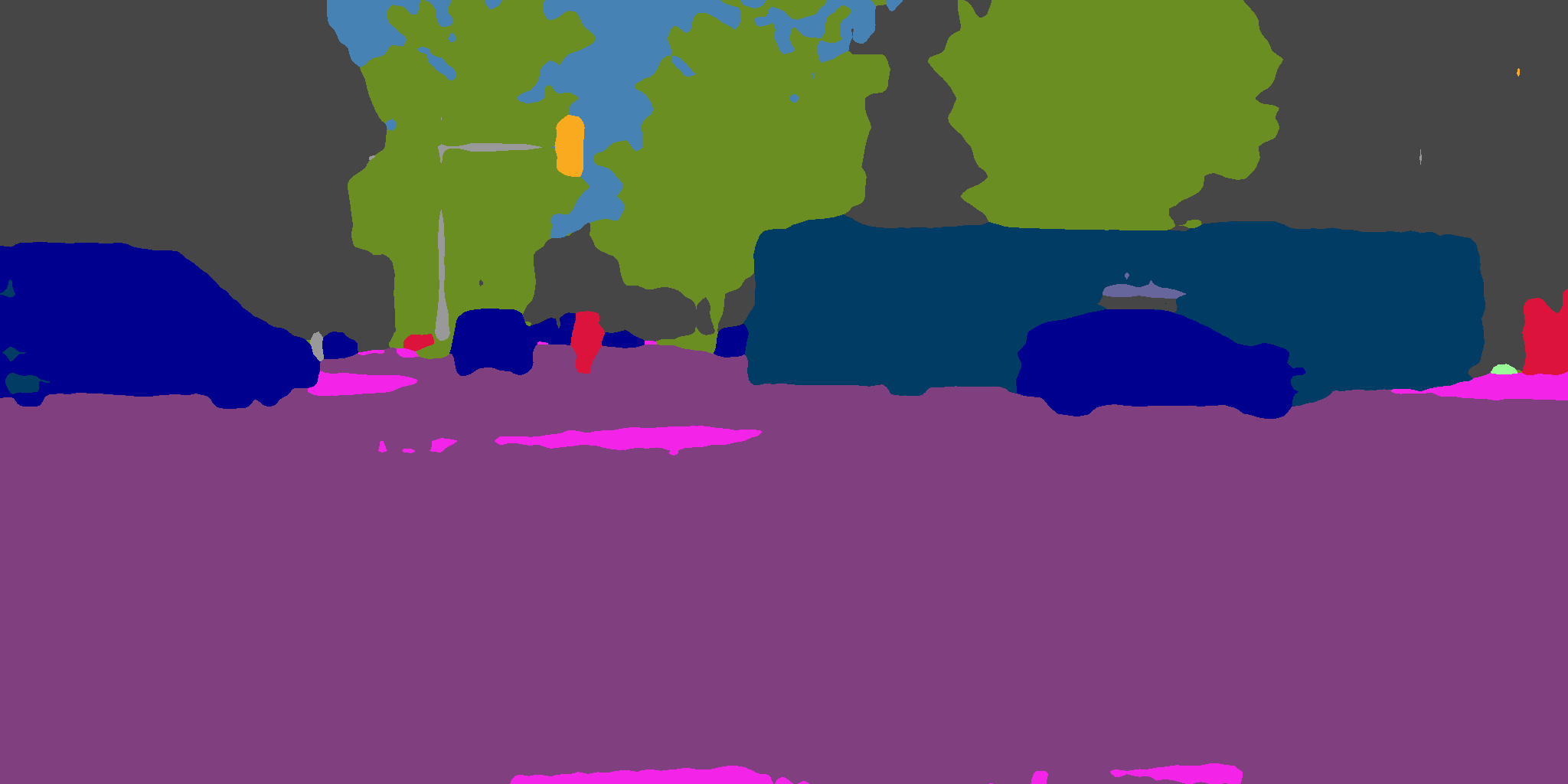}  & \pic{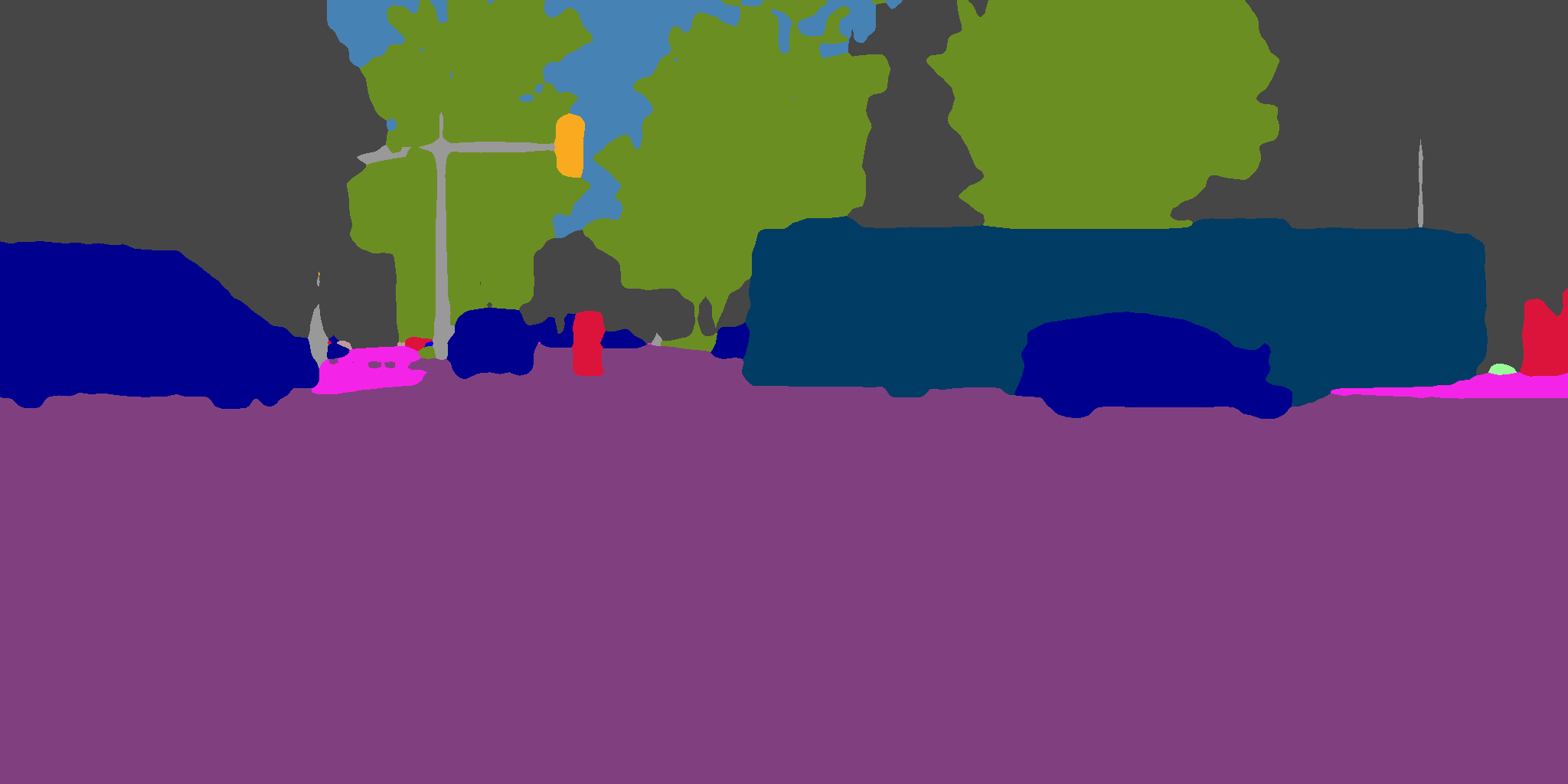} \\
        \pic{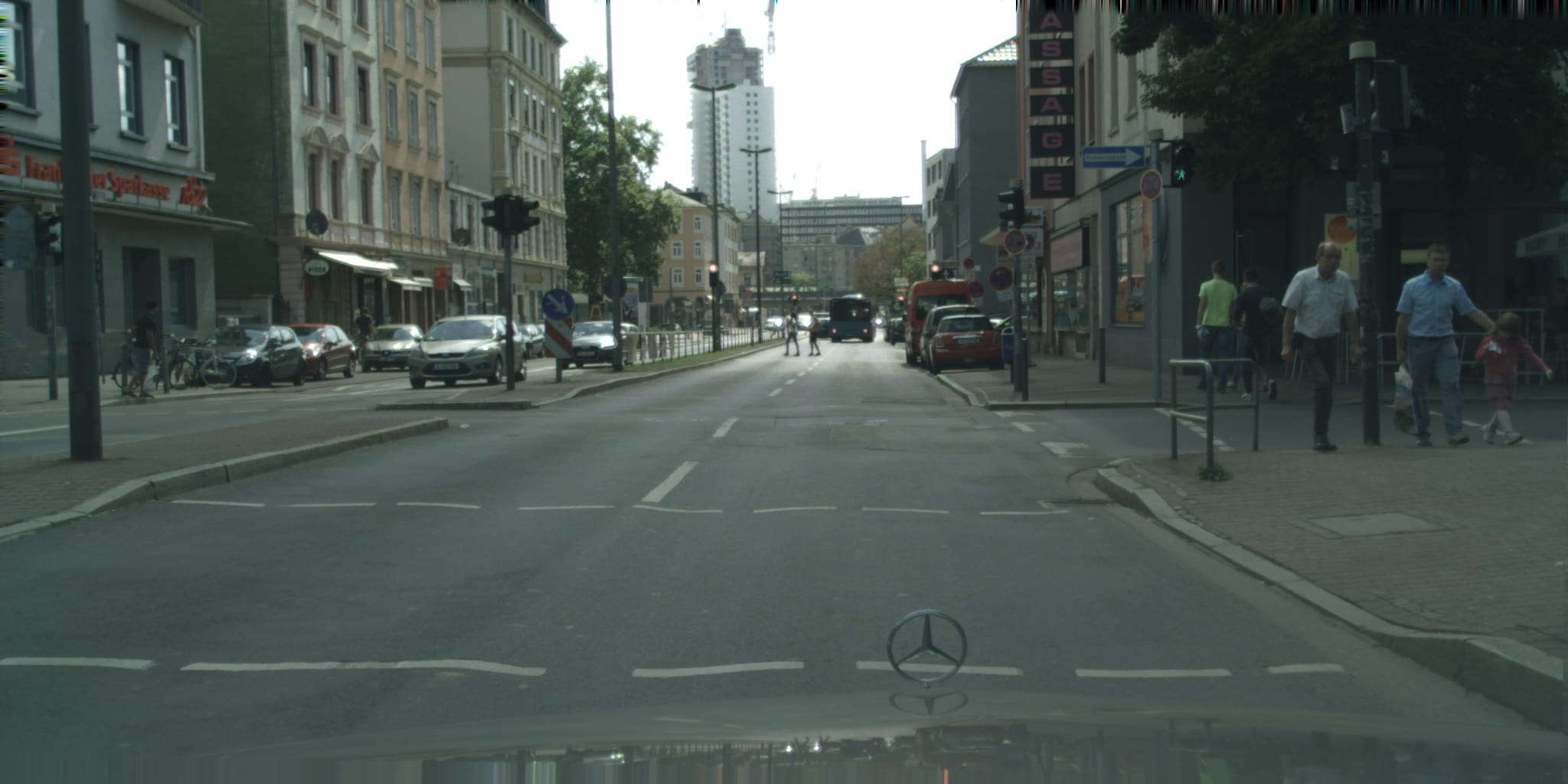} & \pic{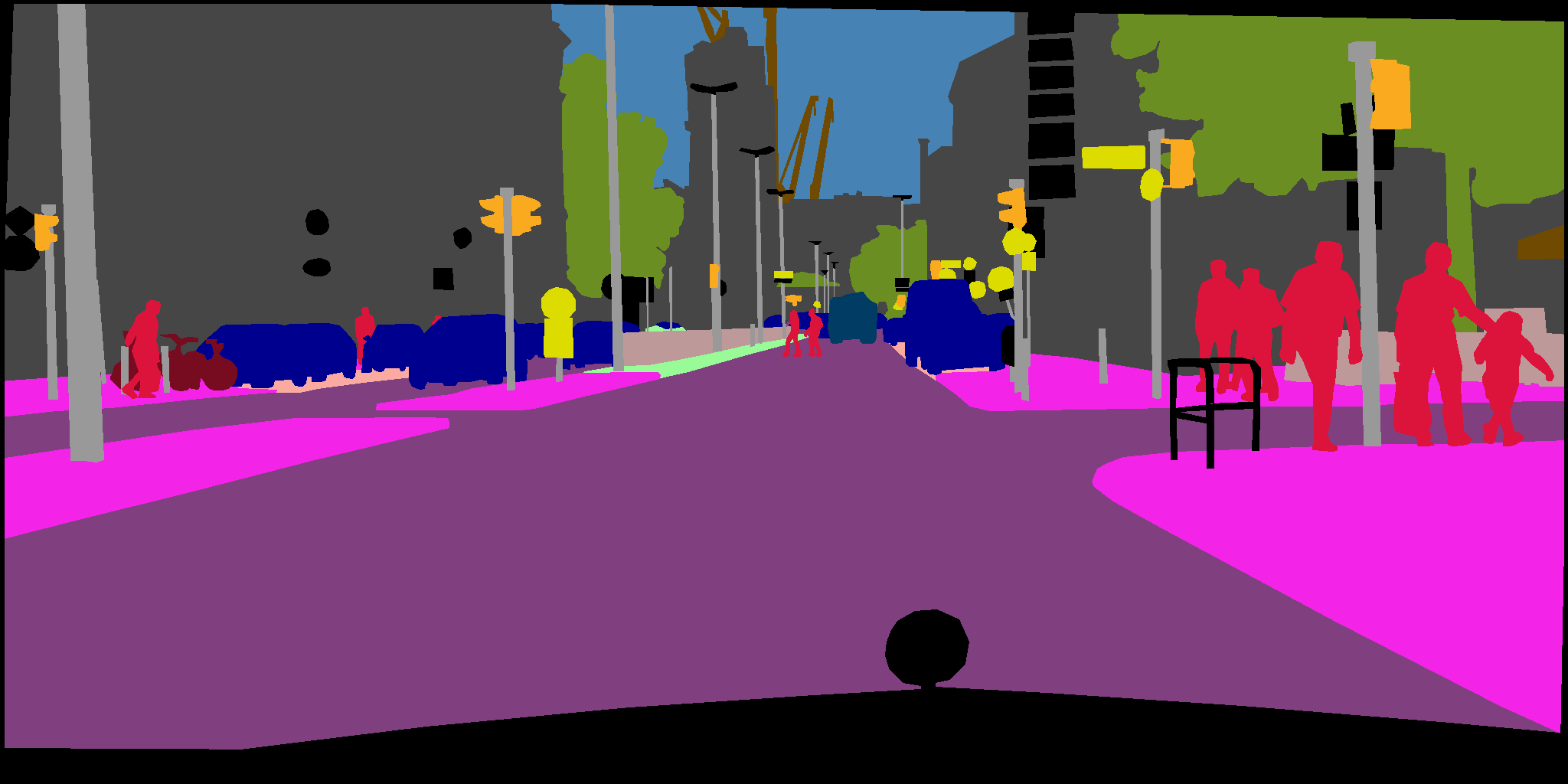}  & \pic{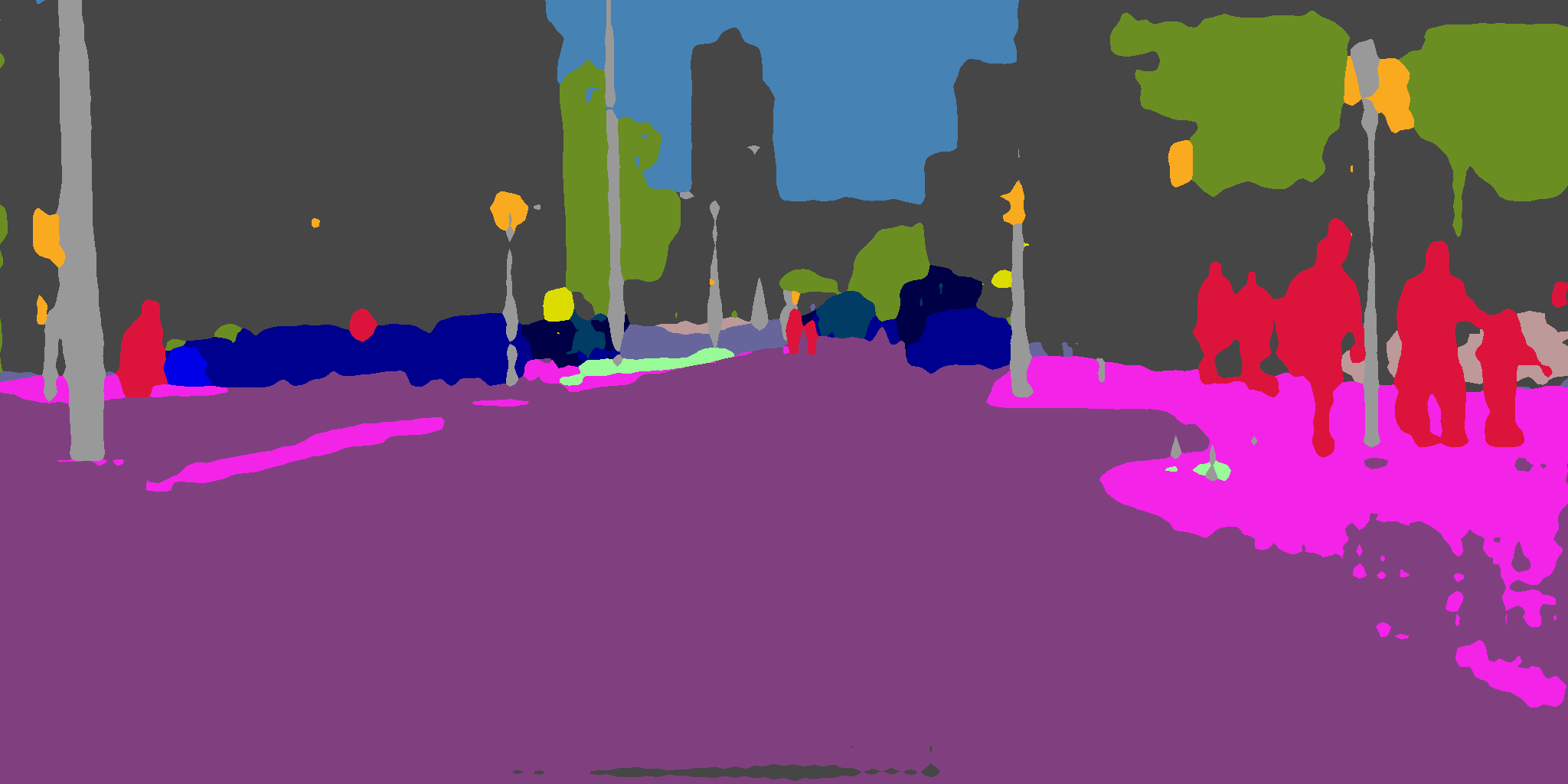}  & \pic{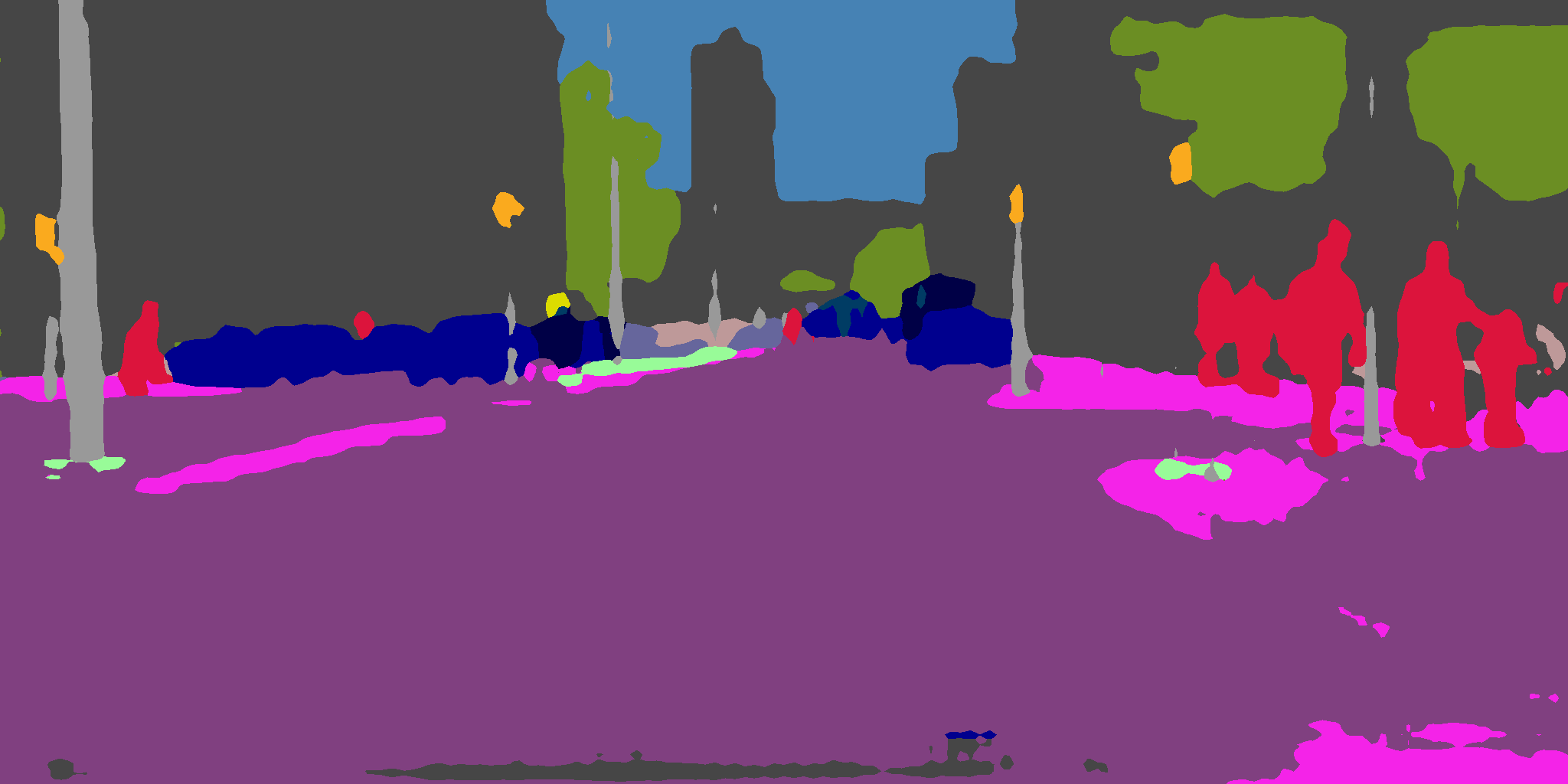}  & \pic{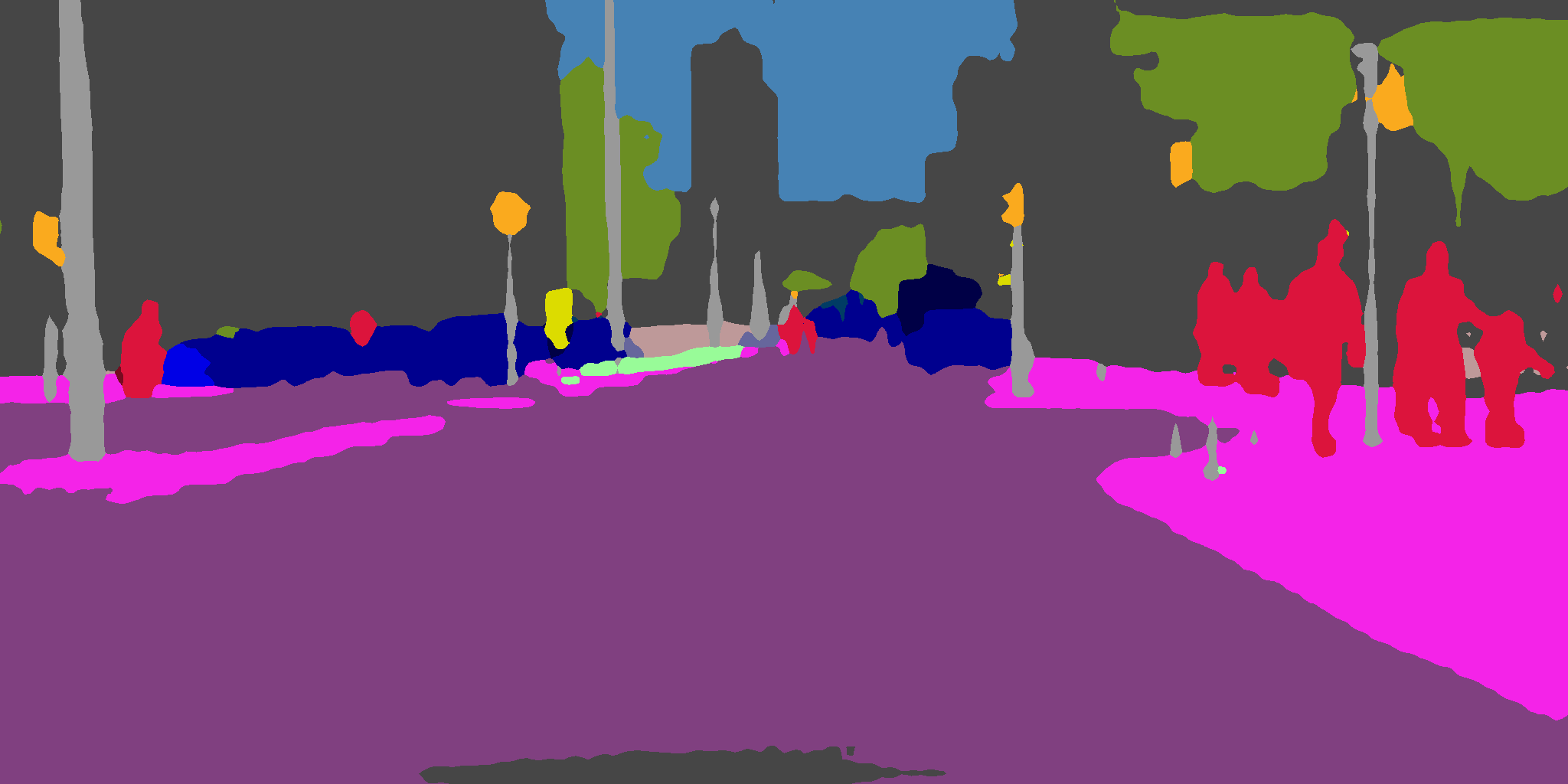} \\
        image & ground truth & pretrain & ensemble & tri-learning \\
    \end{tabular}
    \caption{Visualization of the segmentation results for tri-learning architecture and implicit pseudo supervision. ``pretrain'' represents the best model in the pretrain stage. ``ensemble'' represent the ensemble of three networks. ``tri-learning'' represents the best model in the training stage.  }
    \label{tab:display_in}
\end{figure*}
\subsection{Settings}
\textbf{Network Architecture.} In the experiments, We use ResNet-101\cite{He2015} as the shared backbone network, which is widely adopted in the adaptive semantic segmentation tasks. Then we design three segmentation networks with different structures and depths. The first segmentation network $F_1$ is the same as the ResNet-101 last layer, segmentation network $F_2$ removes one blocks from $F_1$, and segmentation network $F_3$ substitutes the 3x3 kernel of two blocks to 1x5 and 5x1 kernel respectively. For the three segmentation networks, the astrous spatial pyramid pooling(ASPP)\cite{Chen2016} is adopted as the last layer as classifier.

\textbf{Datasets.} We evaluate the proposed EPS-UDA method for semantic segmentation on the popular synthetic-to-real adaptation tasks: (a) GTA5\cite{Richter2016} $\to$ Cityscapes\cite{Cordts2016} (b) SYNTHIA\cite{Ros2016} $\to$ Cityscapes\cite{Cordts2016}. The GTA5 dataset has 24,966 images that are rendered from the GTA5 games and has the same index-mapping with the Cityscapes dataset. In the experiments, we divide GTA5 dataset into two parts: 24,466 images for training and 500 images for validation. The SYNTHTIA dataset has 9,400 images. We randomly select 8,900 images for training, and the remaining 500 images are regarded as the validation dataset. We only use the same categories because the SYNTHIA dataset has different label mapping with the target dataset Cityscapes. Specifically, 16 categories are selected from the original 19 categories. Following the setting of target dataset in \cite{Tsai2018}, we use 2,975 for training and 500 images for testing to evaluate the effectiveness of the proposed model with mean Intersection over Union (mIoU) and Intersection over Union (IoU) of per-class.


\textbf{Implementation Details.} Our experiments are implemented on Pytorch and ran on one GeForce RTX 2080 with 11G maximum memory. Following \cite{Zheng2020a}, the input images are firstly resized to (1280, 640) jittering from [0.8, 1.2] and then randomly cropped to (640, 360). Horizontal flipping is applied with the possibility of 50\%. The batch size is 2. For the learning rate, we follow the settings in \cite{Tsai2018}. The learning rate for the segmentation networks is set to $2.5e^{-4}$ and $1e^{-4}$ for the discriminator. The weight of $\mathcal{L}_{adv}$ is set to $2e^{-5}$. As for the number of saved feature centroids, $n_B$ equals to $20$, $n_R$ equals to $200$ and $M$ equals to 8. In the proposed method, a pretrained strategy is adopted to initialize the parameters. In the pretrained strategy, $\lambda_{sfa}$ are set to 0.1, $\alpha$ is set to 0 and $\lambda_{aae}$ is set to 1. The weight of $\mathcal{L}_{adv}$ is set to $2e^{-5}$. 
Then the model is refined with the best pretrained model. $\lambda_{sfa}$ a are set to 0.02 and 0.01 for the SYNTHIA dataset and GTA5 dataset. $\lambda_{aae}$ is set to 3 and 1 for the SYNTHIA dataset and GTA5 dataset.

\subsection{Experimental Results}
GTA5 $\to$ Cityscapes and SYNTHIA $\to$ Cityscapes are two popular tasks to verify the effectiveness of the unsupervised domain adaptation method for semantic segmentation. We reported the experimental results on GTA5 $\to$ Cityscapes in Table \ref{tab:gta5} and the experimental results on SYNTHIA $\to$ Cityscapes in Table \ref{tab:synthia}.


ProDA\cite{Zhang2021} ranks first in both two tasks and has two stages. The method in ProDA,  including prototypical pseudo labels and strong augmented view as the first stage, achieves 53.7(denoted as ProDA$^{\dagger}$ in Table \ref{tab:gta5}) on GTA5 $\to$ Cityscapes, which is slightly lower than 54.1(our proposed method). Then ProDA transfers knowledge from the best model in the first stage to a student model in a self-supervised manner, which achieves a very high accuracy in the second stage. There is no data for ProDA$^{\dagger}$ on SYNTHIA $\to$ Cityscapes, so we can not list the IoU in Table \ref{tab:synthia}. Because the self-supervised process of the second stage only involves the teacher-student model rather than prototype alignment or multi-view, the result in the first stage still proves the considerable improvement of our proposed method. 

Meta\cite{Guo2021} exploits the covariance of feature map to evaluate the adaptation ability of each class and re-weights the pseudo labels for each class. But this method can not evaluate the adaptation ability for each pixel, the pseudo labels after re-weighting are still less accurate, causing much lower mIoU.

MRnet\cite{Zheng2020a}, another multi-view method, achieves 50.3 on GTA5 $\to$ Cityscapes. It estimates the uncertainty of the pseudo labels by the variance of two views to rectify the pseudo labels. Because one of the views has a very weak constraint to produce confident pseudo labels, the final prediction is not satisfying.

IAST\cite{Mei2020}, a classic self-training method, has already achieved 52.2. It makes full use of probability distribution in one view and maintains adaptive confidence thresholds for each class. Lack of the more reliable method than probability threshold, this method still suffers the confirmation bias in pseudo labels.

Considering other state-of-the-art methods, such as Pix, Dacs, PIT, all these results show that the proposed method is promising by utilizing the pseudo labels for adaptive semantic segmentation task.
  

\begin{table}
\centering
  \caption{Ablation study(GTA5 $\to$ Cityscapes). }
  \label{tab:ablation}
  \begin{tabular}{ccccc}
    \hline
     Discriminator & Tri-learning & SFA & AAE &  mIoU\\
    \hline
              & & & & 36.6 \\
      $\surd$ & & & & 41.3 \\
      $\surd$ & & $\surd$ & & 46.2 \\
      $\surd$ & ensemble & & & 46.1 \\
      $\surd$ & $\surd$ & back &  & 47.0 \\
      $\surd$ & $\surd$ & fore & & 49.5 \\
      $\surd$ & ensemble & $\surd$ & & 48.7 \\
      $\surd$ & $\surd$ & $\surd$ & & 50.5\\
  \hline

    $\surd$ &  $\surd$  & $\surd$ &          & 50.9 \\
    $\surd$ &  $\surd$  &         & $\surd$  & 53.4 \\
    $\surd$ &  ensemble &         & $\surd$  & 52.0 \\
    $\surd$ &  $\surd$  & $\surd$ & $\surd$  & 54.1 \\
  \hline
\end{tabular}
\end{table}

\subsection{Ablation Study}
In this section, we verify the effectiveness of different parts in the proposed method, including SFA, AAE, the tri-learning architecture and discriminator in Table \ref{tab:ablation}. In the pretrain stage, the original model without target data gets 36.1, the same as the other method does. After added the discriminator and target data as the Adapt\cite{Saito2018} does for single layer, the mIoU gets 41.4. Then the best model in pretrain stage generates the fixed pseudo labels for all target training images. In the training process, the best model gets 50.9 with these fixed pseudo labels.

\textbf{Influence of Tri-learning architecture:}  When only applying the ensemble of three segmentation networks, the mIoU has a large leap to 46.1. Compared with the ensemble of three networks, applying tri-learning architecture for SFA gains +1.8 and for AAE gains +1.4. These gains prove the efficiency of the strategy that two networks produce pseudo labels for the third network. 

\textbf{Influence of SFA:} We takes different strategies for background and foreground categories respectively in SFA. The combined strategy gains +2.5 and +1.0 than the single strategy applied to all classes respectively. These gains prove that the efficiency of different strategies. The complex strategy that considers the relationship between different classes may undermine the inherent semantic feature structure for background categories. The simple strategy that only consider the same class may have poor discrimination between foreground categories. The overall SFA has gained +4.4 in total.

\textbf{Influence of AAE:} After applying the AAE, the model gains +3.2 in total. The reason may be that the different views cannot be ensured with high probability although they provide the same prediction. AAE can force such pixels to be aligned well. 

\section{Conclusion}
In this paper, we propose an implicit pseudo supervision for unsupervised domain adaptation for semantic segmentation. This supervision is based on a tri-learning architecture, which has three segmentation networks and each two networks generate reliable pseudo labels for the third network to keep diversity without regularization. 
Implicit pseudo supervision includes SFA and AAE. Both two methods utilize the pseudo labels implicitly.
SFA attempts to align the semantic feature centroids conditionally for background and foreground categories. 
AAE measures how much a pseudo-labelled pixel can improve the model and rectifies the pseudo labels for each network to provide target-specific knowledge. 
The proposed method is verified on the popular and benchmark segmentation tasks, and  outperforms several state-of-the-art methods considerably.

\section*{Acknowledgment}

Sincere gratitude to anonymous reviewers for careful work and considerate suggestions. 

\bibliographystyle{IEEEtran}
\bibliography{conference_101719}


\end{document}